\documentclass[11pt]{article}
\usepackage[top=1in, bottom=1in, left=1in, right=1in, letterpaper]{geometry}

\usepackage[T1]{fontenc}

\usepackage[sc,osf]{mathpazo}   
\usepackage[euler-digits,small]{eulervm}

\usepackage{booktabs}       
\usepackage{amsmath, amssymb, amsfonts, bm}
\usepackage{amsthm}
\usepackage{mathtools}
\usepackage{nicefrac}       
\usepackage{microtype}      
\usepackage{subcaption}
\usepackage[usenames, dvipsnames]{color}
\usepackage{floatrow}
\usepackage{graphicx}
\usepackage{algorithm}
\usepackage{algorithmicx}
\usepackage[noend]{algpseudocode}
\usepackage{amssymb}
\usepackage{multirow,bigdelim}
\usepackage{textcomp}
\usepackage[title]{appendix}
\usepackage[percent]{overpic}
\usepackage{floatrow}

\usepackage{cite}
\usepackage[hyphens]{url}
\usepackage[colorlinks=true,citecolor=blue,linkcolor=blue,bookmarksnumbered=true,hyperfootnotes=true,breaklinks]{hyperref}
\usepackage[nameinlink]{cleveref}

\theoremstyle{remark}

\def\Fig#1{Fig.~\ref{#1}}

\DeclareMathOperator{\softplus}{softplus}
\DeclareMathOperator{\softmax}{softmax}
\DeclareMathOperator{\edit}{edit}

\DeclareMathOperator{\convolution}{convolution}
\DeclareMathOperator{\attn}{attn\_control}

\newcommand{\seq}[1]{\mathbf{#1}}
\newcommand{\xx}{\seq{x}}
\newcommand{\yy}{\seq{y}}

\newcommand{\otag}{\underbracket[0.7pt][0.7pt]}
\newcommand{\odum}[1]{\otag{\hspace{#1}}}

\newcommand{\rot}{{\scriptscriptstyle \circlearrowleft}}
\newcommand\Vtextvisiblespace[1][.3em]{%
  \mbox{\kern.06em\vrule height.3ex}%
  \vbox{\hrule width#1}%
  \hbox{\vrule height.3ex}}
\newcommand{\dummy}{\hspace{1.2pt} \text{\Vtextvisiblespace[0.7em]}\hspace{1.2pt}}

\newcommand{\COMMENT}[1]{}

\title{Learning to Remember, Forget, and Ignore using Attention Control in Memory}

\author{%
	T.S. Jayram\thanks{Equal contribution} 
	\and Younes Bouhadjar\footnotemark[1]
	\and Ryan L. McAvoy 
	\and Tomasz Kornuta 
	\and Alexis Asseman 
	\and Kamil Rocki 
	\and Ahmet S. Ozcan \\
	IBM Research AI, Almaden Research Center, San Jose, USA\thanks{%
		Contacts: \texttt{jayram@us.ibm.com}, \texttt{younes.bouhadjy@gmail.com}, 
		\texttt{mcavoy@us.ibm.com},
		\texttt{alexis.asseman@ibm.com}, 
		\texttt{tkornut@us.ibm.com},
		\texttt{kmrocki@us.ibm.com}, 
		\texttt{asozcan@us.ibm.com}}
}
\date{}

\begin{document}

\maketitle

\begin{abstract}
 Typical neural networks with external memory do not effectively separate capacity for episodic and working memory as is required for reasoning in humans. Applying knowledge gained from psychological studies, we designed a new model called Differentiable Working Memory (DWM) in order to specifically emulate human working memory. As it shows the same functional characteristics as working memory, it robustly learns psychology inspired tasks and converges faster than comparable state-of-the-art models. Moreover, the DWM model successfully generalizes to sequences two orders of magnitude longer than the ones used in training. Our in-depth analysis shows that the behavior of DWM is interpretable and that it learns to have fine control over memory, allowing it to retain, ignore or forget information based on its relevance.	
\end{abstract}

\section{Introduction}

Keeping information in mind after it is no longer present in the environment is critical for all higher cognitive behaviors. \emph{Working memory} (WM) is the term used for this ability, which is distinct from the storage of vast amount of information in long-term memory~\cite{baddeley2003, oberauer2009}.  The two main distinguishing characteristics of WM are the limited capacity (3-5 items)~\cite{conway2005} and temporary retention (secs-minutes).  Hence, WM is not a storage per se, but a mental workspace utilized during planning, reasoning and solving problems. Most psychologists differentiate WM from ``short-term'' memory because it can involve the manipulation of information rather than being a passive storage \cite{cowan2017}.
Along the same lines, Engle et al.~\cite{engle1999}
argued that WM is \emph{all about the capacity for controlled, sustained attention in the face of interference or distraction}.
Attention-control is a fundamental component of the WM system and probably the main limiting factor for capacity~\cite{conway1994working,engle2004executive}.  Consequently, the inability to effectively
parallel process two-attention demanding tasks limits our multitasking performance severely. 

Over the past several decades psychologists have developed tests to measure the individual differences in WM capacity and better understand the underlying mechanisms.  These tests have been carefully crafted to focus on the specific aspects of WM such as task-driven attention control, interference and capacity limits~\cite{oberauer2017}. The best known and successfully applied class of tasks for measuring WM capacity is the ``complex span'' paradigm.  The challenge presented by complex span tasks is recalling the list of items, despite being distracted by the processing task. Studies show that individuals with high WM capacity are less likely to store irrelevant 
distractors~\cite{vogel2005neural} and they are better at retaining task-relevant information~\cite{maxcey2013strategic}. 
Developing task-driven strategies for cognitive control are essential for the effective use of WM.


The power of maintaining information over time has also been recognized by the AI community.  
Starting with the basic recurrent neural network architectures~\cite{elman1990finding,hopfield1986computing}
followed by the introduction of gating mechanisms~\cite{hochreiter1997long}, 
the research has recently moved onto more complex architectures with memories~\cite{graves14,joulin2015inferring,weston2015memory,graves2016hybrid,santoro2016meta,gulcehre2017memory}.
These models are typically applied to tasks (e.g. associative recall, bAbI QA~\cite{westonBCM15}) that require a complex mixture of long-term memory (episodic and semantic) and working memory. In the human brain, these kinds of memory systems are distinct: working memory is instantiated in multiple interconnected areas with the prefrontal cortex playing a major role \cite{Constantinidis2016}, whereas for episodic memory the hippocampus is the critical structure \cite{Fortin2002}.
Studying these mechanisms separately is necessary to disentangle the contributions of each memory system and develop a detailed understanding of human intelligence.

In this work, we introduce a Differentiable Working Memory (DWM) model that more closely mimics the functional behavior of human WM. We also present a battery of tasks adopted from the cognitive psychology literature that allow us to elucidate the working memory behavior of a neural network directly. In contrast to prior work on neural nets with external memory, our starting point is the functional attributes of WM and the tasks that primarily tests those.

The main contributions of this work are:
\begin{itemize}
\item A differentiable WM model that is easy to train and accurately generalizes to sequences two orders of magnitude longer than the training data.
\item A new attention control mechanism in memory, which can deal with interference.
\item Algorithmic working memory tasks adapted from the cognitive psychology.
\item An examination of strategies developed for handling memory scarcity.
\end{itemize}

\section{Methods}
\subsection{Psychometric tasks for working memory}
\begin{figure}[!b]
  \centering
  \includegraphics[width=\textwidth]{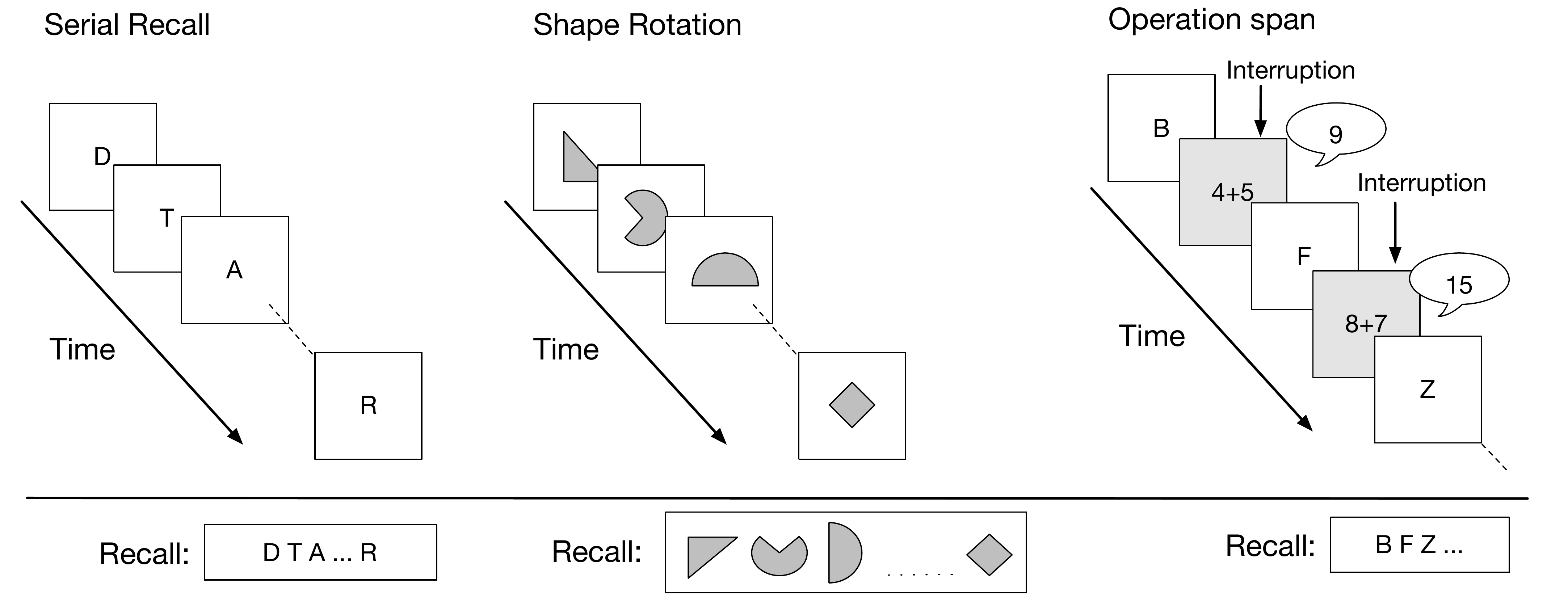}
  \caption{Exemplary tasks for testing the performance of human working memory} \label{fig:human_wm_tasks}
\end{figure}

Over last several decades cognitive psychologists have developed many tasks to measure the performance of human WM (See \Fig{fig:human_wm_tasks} for examples).  These tasks are mainly sequential and typically divided into verbal and visuospatial domains. One of the fundamental goals of these psychometric tests is to measure individual differences and correlate to performance in reasoning and fluid intelligence. In this regard, WM capacity, retention and the ability to switch between tasks are the key predictors.

\begin{figure}[!t]
  \centering
  \includegraphics[width=\textwidth]{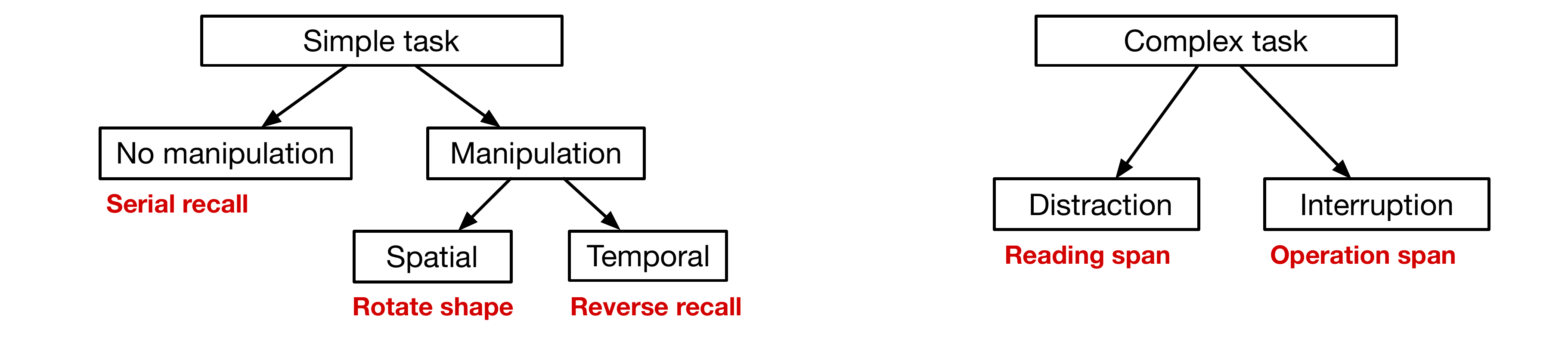}
  \caption{Taxonomy of working memory tasks}
  \label{fig:taxonomy}
\end{figure}

Given the large number of WM tasks in the psychology literature and various categorizations by different researchers, we built a taxonomy of tasks (\Fig{fig:taxonomy}) and carefully selected tasks that seem to be the most representative for a given category. First order categorization is based on the number and complexity of tasks.
For simple tasks, the presence of data manipulation is the next level sub-category. Serial recall is a prime example for a simple task without manipulation.  Other WM tasks may require the manipulation of the memory content, which could be divided in spatial and temporal domains.  The complex WM tasks involve multiple sequential inputs or sub-tasks but not necessarily implies “multi-tasking”.  We follow the framework developed by Clapp and Gazzeley~\cite{clapp2009mechanisms} to distinguish the sources of goal interference, i.e. Distraction (to-be-ignored) and Interruption (i.e. multi-tasking).  For example, Operation Span (\Fig{fig:human_wm_tasks}c) is a dual task because the subjects must attend and process the summation (Interruption) even though they do not need to recall the results afterwards. In Reading Span (1980 version by \cite{daneman1980individual}) subjects read sentences and need to recall the last word of each one.

\subsection{Differentiable Working Memory (DWM)}
\label{sec:attention}
\begin{figure}[!b]
    \centering
    \includegraphics[width=\textwidth]{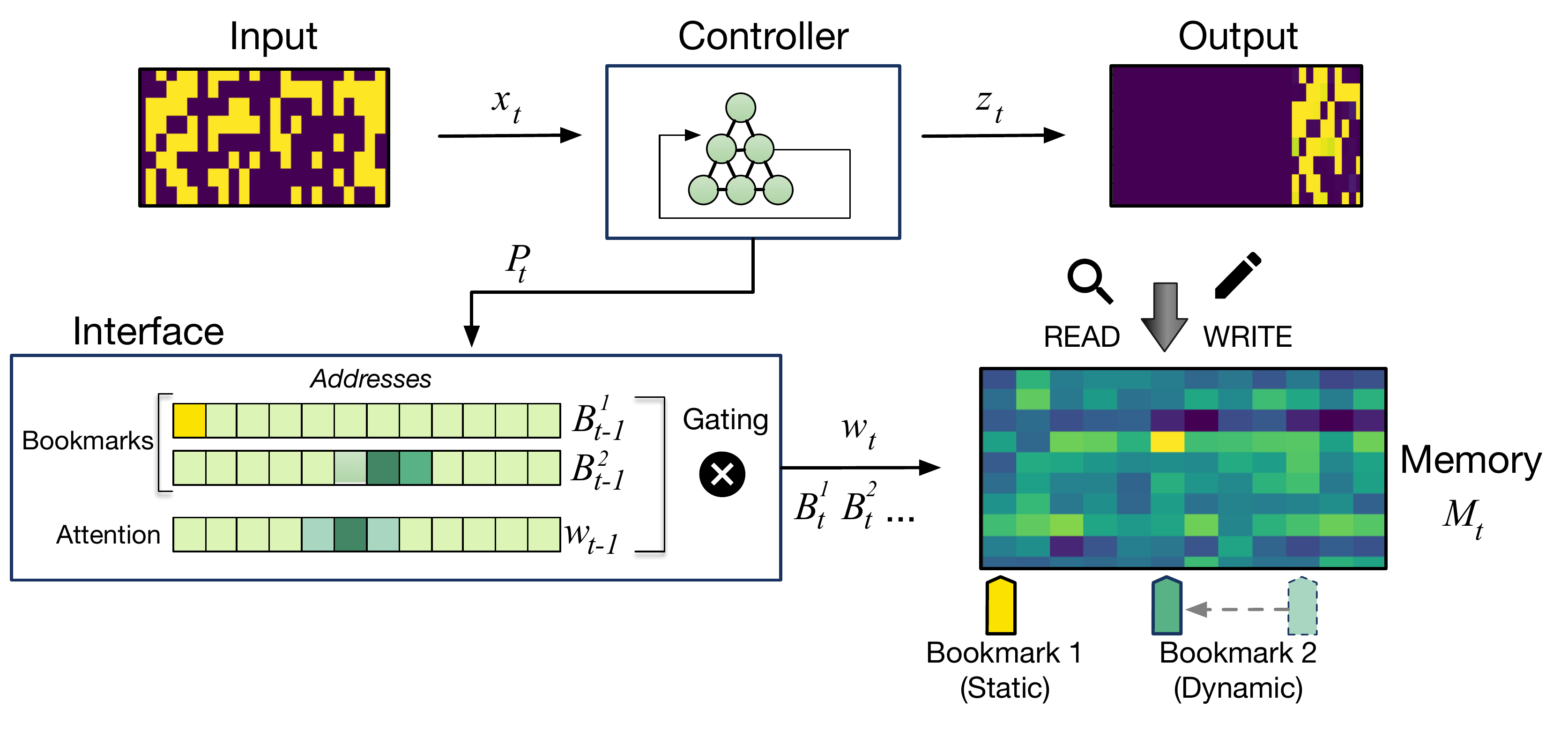}
    \caption{Illustration of the operation of the DWM Model}
    \label{fig:DWM}
\end{figure}

Inspired by human working memory, we designed a Differentiable Working Memory model with the appropriate functional characteristics. We illustrate its operation on \Fig{fig:DWM}. As a neural network with external memory, the DWM has three main components: a \emph{controller}, an \emph{external memory} and an \emph{interface} between the two \cite{zaremba2016learning,zaremba2016learning_icml}. The controller generates parameters that allow the interface to pay \emph{attention} to specific locations in memory and then read and write to them. The procedure is sketched in Algorithm~\ref{algo:DWM}. 
The detailed equations for steps 4--6 can be found in the Supplemental Information.

\paragraph{Attention control.}
For our model to act like human working memory, it must have the same characteristics, the most important of which is that working memory is accessed and written to sequentially~\cite{singh2018working}. 
One condition is that it has to use circular convolution shifting enabling it to shift attention over memory. 
For that purpose we use a mechanism similar to the one used by Neural Turing Machine (NTM)~\cite{graves14}:  
\begin{equation}
w_{t}= \convolution(w^g_{t}, s_t)
\label{eq:convolution}
\end{equation}
where $w^g_t$ and $w_t$ are the vectors of attention weights over cells in memory at time $t$ before and after shifting, and $s_t$ is a shift vector outputted by the controller. The other condition that sequentiality suggests is that the read and write operations should jointly share a single attention so that all access happens in sequential order. As is standard with the linear shifting in memory enhanced neural networks, we also apply a weight sharpening step as detailed in Graves \textit{et al.} \cite{graves14}.

\begin{algorithm}[!t]
	\caption{Operation of the Differentiable Working Memory}
	\label{algo:DWM}
	\begin{algorithmic}[1]
		\State{Initialize the hidden state $h_0$ and memory array $M_0$}
		\State{Initialize the attention vector $w_0$ and
		           bookmark vectors $\{B_0^i : i = 1,2,\dots, N_B\}$}
		\For{$t \in \{1,2, \dots, T \}$} 
			\State{Read from the memory: $r_{t-1} \gets M_{t-1} w_{t-1}$}
			\State{Compute a new controller hidden state and parameters: $h_t$, $P_t \gets \phi(x_t,r_{t-1},h_{t-1})$}
			\State{Write and erase from memory: $M_t \gets \edit(w_{t-1}, P_t, M_{t-1})$}
			\State{Update attention and bookmarks: $w_t, \{B^i_t\}= \attn( w_{t-1}, \{B^i_{t-1}\}, P_t)$}
		\EndFor
	\end{algorithmic}
\end{algorithm}

Working memory also involves some limited model of when information was attended to in the past~\cite{singh2018working}.
This characteristic suggests that human working memory could contain some limited internal record of its past attention. In DWM, this is accomplished by storing \emph{bookmarks} of 
system's attention at previous time steps.
This is recorded in $N_B$ bookmark vectors 
$\{B_t^i : i = 1,2,\dots, N_B\}$ at time $t$.
At each time step, the DWM must decide whether to remember its previous attention $w_{t-1}$ by recording it in a bookmark, as: 
\begin{equation}
B^{i}_t= g^{i}_t w_{t-1} + (1-g^{i}_t)B^{i}_{t-1},
\label{eq:bookmark_update}
\end{equation}
where the gating parameter $g^{i}_t$ is emitted by the controller. Additionally, the DWM keeps one bookmark fixed to the initial attention at time $t=0$ so the model maintains a fixed reference frame. As discussed below, we found even when limiting the bookmark memory to only two bookmarks that we could still solve all tasks.

The DWM must also decide before moving sequentially whether it wishes to return to a previous bookmark. For this purpose we once again use a gating mechanism, this time in a slightly more sophisticated form:
\begin{equation}
w^g_t= \delta^0_t w_{t-1} + \sum_{i=1}^{N_B}\delta^i_t B^i_{t-1},
\label{eq:attention_update}
\end{equation}
where $\delta^i_t, i = 0 \ldots N_B$ are gating parameters emitted by the controller. These gating parameters are scalars, normalized using a softmax function.

The DWM attention control incorporates the presented mechanisms using the combination of equations \eqref{eq:attention_update},\eqref{eq:bookmark_update},\eqref{eq:convolution} in order.  

\subsection{Formal description of tasks}
In order to investigate the capabilities of working memory (WM) such as retention, forgetting and ignoring we introduce a battery of tasks presented in \Cref{tab:tasks}.
These tasks are motivated by, and have a direct correspondence to the categorization shown in~\Fig{fig:taxonomy}, but modified so that they are agnostic to audio/visual processing. 
In addition to the classical psychometric tasks, we introduce
additional tasks that will test the effectiveness of attention control in 
memory (Ignore, Forget and Scratch Pad).

The input to every WM task is a time-indexed stream of items.
At a higher level, we view the input as a  \emph{concatenation} of various subsequences that represent different functional units of processing.
Additionally we use a constant-sized set of special items (called command markers) to both mark the beginning of a subsequence as well as indicate its functional type. Important note is that the system does not know a priori what kind of operation is associated with a given type of marker and must learn that from data.  We ignored markers in Table \ref{tab:tasks} to keep the description simple. Also, note that such markers are also commonly employed in the psychometric tests (e.g. see \cite{McNab2008}).

For all Simple tasks, there is only one type of subsequence, and the 
output will be reproduced from the memory with or without manipulation.
The $|$ sign indicates the delay between input and output of the primary subsequence(s).
In our notation, $x^{\rot}$ represents doing a circular shift of the bit representation of element $x$  by half the number of bits.
The Complex tasks may involve a secondary set of subsequences, which may or may not require immediate output as indicated in the Forget and Operation Span tasks.

\begin{table}[hbtp] 
\def\arraystretch{1.2}
\begin{center}
	\begin{tabular}{l@{\extracolsep{5pt}}l@{\extracolsep{5pt}}r@{\extracolsep{5pt}}r@{\extracolsep{1pt}}l@{\extracolsep{2pt}}l}
		\toprule
		 \multicolumn{2}{c}{Task} & \multicolumn{3}{c}{(I)input/(O)output sequences} & Notes \\ 
		\midrule
	\multirow{6}{*}{\rotatebox[origin=c]{90}{Simple}} &\multirow{2}{*}{Serial Recall}& I: & 
		  $x_1 x_2 \dots x_n$ & 
		  $| \odum{0.35cm} \odum{0.35cm} \dots \odum{0.35cm}$  &  
		  Store the input and recall it\\
		&  & O: & $\odum{0.4cm} \odum{0.4cm} \dots \odum{0.4cm}$ & 
		$|$ $ x_1 x_2 \dots x_n$ & in the same order\\
		\cline{2-6}
		& \multirow{2}{*}{Reverse Recall} & I: & 
		$x_1 x_2 \dots x_n$ & 
		 $| \odum{0.4cm} \odum{0.7cm} \dots \odum{0.4cm}$  &  
		   Store the input and recall it \\
		&  & O:& $\odum{0.4cm} \odum{0.4cm} \dots \odum{0.4cm}$ & 
		$|$ $x_n x_{n-1} \dots x_1$ & in the reversed order\\
		\cline{2-6} 
		& \multirow{2}{*}{Rotate Shape}  & I: & 
		$x_1 x_2 \dots x_n$ & 
		$| \odum{0.4cm} \odum{0.4cm} \dots \odum{0.4cm}$  &  
		Rotate each element of sequence \\
		&  & O: & $\odum{0.4cm} \odum{0.4cm} \dots \odum{0.4cm}$ & 
		$|$ $ x_1^{\rot} x_2^{\rot} \dots x_n^{\rot}$ &  \\
		\midrule
		\multirow{10}{*}{\rotatebox[origin=c]{90}{Complex}} & \multirow{2}{*}{Reading Span} & I: & $\xx_1 \dots \xx_k $ & $| \odum{0.35cm} \dots \odum{0.35cm}$  &  Recall $z_i$ being the last  \\ 
		&  & O: & $\odum{0.40cm} \dots \odum{0.40cm} $ & $|$ $ z_1 \dots z_k$  & element of $\xx_i$  \\[1pt]
		\cline{2-6} 
		& \multirow{2}{*}{Forget}  & I: & $\xx_1 \yy_1 \odum{0.4cm} \dots \xx_k \yy_k \odum{0.4cm}  $ & $| \odum{0.4cm} \dots \odum{0.4cm} $ & Recall  every $\yy_i$ immediately, \\
		& & O: & $\odum{0.35cm} \odum{0.35cm} \yy_1 \dots \odum{0.35cm} \odum{0.35cm} \yy_k  $ & $|$ $ \xx_1 \dots \xx_k$ & recall all $\xx_i$ in the same order \\
		\cline{2-6}
		& \multirow{2}{*}{Operation Span}  & I: & $\xx_1 y_1 \odum{0.45cm} \dots \xx_k y_k \odum{0.45cm} $ & $| \odum{0.4cm} \dots \odum{0.4cm} $ & Rotate $y_i$ immediately, \\
		&  & O: & $ \odum{0.35cm} \odum{0.35cm} y_1^{\rot} \dots \odum{0.35cm} \odum{0.35cm} y_k^{\rot} $ & $|$ $ \xx_1 \dots {\xx_k}$ & recall all $\xx_i$  in the same order\\
		\cline{2-6}
		& \multirow{2}{*}{Scratch Pad}  & I: & $\xx_1 \dots \xx_k  $ & $|  \odum{0.4cm}$  & Return only the last $\xx_k$ \\
		&  & O: & $ \odum{0.4cm} \dots \odum{0.4cm} $ & $|$ $ {\xx_k}$ & \\
		\cline{2-6}
		& \multirow{2}{*}{Ignore}   & I: & $\xx_1 \yy_1 \dots \xx_k \yy_k $ & $|$ $  \odum{0.4cm} \dots \odum{0.4cm}$ & Ignore $\yy_i$, recall $\xx_i$ \\
		&  & O: & $ \odum{0.35cm} \odum{0.35cm} \dots \odum{0.35cm} \odum{0.35cm} $ & $|$ $ \xx_1 \dots \xx_k $ & \\
		\bottomrule 
	\end{tabular} 
\end{center}
\caption{The Working Memory Tasks used in our experiments}
\label{tab:tasks}
\def\arraystretch{1}
\end{table}

\section{Experimental results}

We evaluated the performance of DWM on the proposed tasks and compared it to two models: LSTM (Long Short-Term Memory)~\cite{hochreiter1997long}, considered as a classical baseline for sequential problems, and DNC (Differentiable Neural Computer)~\cite{graves2016hybrid} being the state-of-the-art memory augmented neural model. 
In our implementation we used the Pytorch framework~\cite{paszke2017automatic}.
During training we used the Adam (Adaptive Momentum) optimizer~\cite{kingma2014adam} and (average) binary cross-entropy as the loss function.
We apply early stopping based on validation loss ($10^{-4}$).
Additionally we terminate training when the number of training episodes reach 100,000, where a single episode involves processing a batch of sequences.
The size of batch was a hyper-parameter that was tuned for each model along with training rate using validation loss. The exact parameters describing each tasks can be found in the Supplemental Information. 

It is well-known that neural networks augmented by external memory learn algorithms to process tasks rather than memorizing data~\cite{graves14,graves2016hybrid}. Therefore, to determine the robustness of our models we train them on sequences of lengths of order 10 and then validate and test them on sequences of size 100 and 1000, respectively. As seen in \Cref{tab:results}, the DWM, the LSTM and the DNC are all capable of learning each of the tasks for sequences of order 10. To train on these tasks, the DWM required on average the fewest number of sequences with only Forget and Operation Span requiring more than 10000 batches of size 16 to converge. The DNC required the most number of sequences with only Serial Recall, Rotate Shape and Ignore converging with fewer than 20000 batches of size 16 and the rest requiring between 20000 to 90000 batches.

Only the DWM was able to successfully generalize to sequences of size 1000 even with only 1066 trainable parameters. The DNC with its 4,792 trainable parameters generalized to sequences of length 100 just on Serial Recall, Scratch Pad and Rotate Shape and did not fully generalize on sequences of length 1000 for any of the tasks. Despite that the LSTM possessed over 5 million trainable parameters it was only able to generalize to sequences of length 100 on Serial Recall.

The DWM performance on these tasks indicates that it is actually acting as a form of artificial working memory. Human working memory's operation is straightforward compared to episodic memory and therefore it is more effective at solving the relatively simple working memory tasks. We hypothesize that the DNC, which was explicitly designed to mimic an episodic memory \cite{graves2016hybrid}, did not learn these simple working memory tasks as effectively because the complexity required for true episodic remembrance interfered with learning tasks that do not require episodic memory.




\begin{table}[!t] 
\begin{center}
\begin{tabular}{l@{\extracolsep{5pt}}c@{\extracolsep{5pt}}c@{\extracolsep{5pt}}c@{\extracolsep{5pt}}c@{\extracolsep{5pt}}c@{\extracolsep{5pt}}c@{\extracolsep{5pt}}c@{\extracolsep{5pt}}c@{\extracolsep{5pt}}c@{\extracolsep{5pt}}}
\toprule
\multirow{3}{*}{Task} & \multicolumn{3}{c}{Best Train Accuracy} & \multicolumn{3}{c}{Validation Accuracy} & \multicolumn{3}{c}{Test Accuracy} \\
 & \multicolumn{3}{c}{Seq. Size 10 [\%]} & \multicolumn{3}{c}{Seq. Size 100 [\%]} & \multicolumn{3}{c}{ Seq. Size 1000 [\%]} \\
\cmidrule(l){2-4} \cmidrule(l){5-7}  \cmidrule(l){8-10}
& LSTM & DNC & DWM & LSTM & DNC & DWM & LSTM & DNC & DWM \\
\midrule
Serial Recall   & 100 & 100 & 100 & 100 & 100 & 100 & 50.20 & 64.64 & 100 \\
Reverse Recall & 100 & 100 & 100 & 52.96 & 50.62 & 100 & 50.38 & 50.15 & 99.76 \\
Rotate Shape   & 100 & 100 & 100 & 52.17 & 100 & 100 & 50.20 & 60..91 & 100 \\
Reading Span   & 100 & 100 & 100 & 50.90 & 53.36 & 100 & 50.44 & 49.04 & 91.88 \\
Forget         & 100 & 100 & 100 & 55.90 & 69.36 & 98.92 & 50.45 & 49.94 & 94.11 \\
Operation Span  & 100 & 100 & 100 & 58.16 & 79.22 & 99.95 & 51.26 & 53.61 & 99.64 \\
Scratch Pad    & 100 & 100 & 100 & 71.28 & 100 & 100 & 70.02 & 74.97 & 100 \\
Ignore       & 100 & 100 & 100 & 56.13 & 69.32 & 100 & 50.89 & 49.99 & 90.05 \\
\bottomrule
\end{tabular}
\end{center}
\caption{Summary of experimental results. The first column is the average of the best accuracy achieved during training for each run. The second column is the average of the best validation accuracy for each run. The third column is the average of the test accuracy for the model parameters with the best validation accuracy. For the DNC and LSTM, the validation loss did not reached threshold for some tasks (i.e. training was stopped at 100,000 episodes), in which cases we decided to include the best scores of single best (but diverged) models}
\label{tab:results}
\end{table}

\subsection{Analysis of strategies for solving tasks}

In contrast to long-term memory, working memory has an extremely small capacity. Therefore, dealing with interference (e.g. distractions) is a major challenge for memory capacity and attention control. As mentioned earlier, ignoring distractions without encoding them in the memory is arguably the best strategy to minimize memory consumption. On the other hand, for a complex task with an interruption (i.e. multi-tasking), the secondary task cannot be ignored and may require extensive memory usage.  In this case, the best strategy might be to forget (e.g. erase or overwrite) the secondary information as soon as possible in order to maintain sufficient memory capacity for the main task.  Ignoring and forgetting may be good strategies for the efficient use of capacity but they also complicate the attention control and addressing during writing in the memory. 

During the training and testing of all of the tasks reported in Table~\ref{tab:results}, we provided sufficient memory size, so that the system could store all the encoded input items in the memory (if it has chosen to).
However, limitation of the memory size can force the system to develop more memory efficient strategies, thus we decided to investigate that issue further.


\subsubsection{Strategies for the Scratch Pad task}
The goal of the Scratch Pad task is to recall only the last input subsequence. 
Given the DWM mechanisms, we expect two possible strategies for the model to learn in order to solve this task.

The "Expand" strategy exploits the fact that memory can be used in a similar way to a circular buffer, storing each consecutive subsequences one after the other in the memory. In this case the model should write each subsequence, then place the dynamic bookmark (i.e. the one that does not have fixed attention) at the start of given subsequence, and then update the bookmark position to the beginning of the next subsequence. Finally, when the model receives a command marker indicating it needs to recall, it should recall the attention associated with that dynamic bookmark and then retrieve consecutive items one by one using circular convolution. 

The "Overwrite" strategy for the Scratch Pad relies on the fact that when a new subsequence appears, the elements from the previous one can be discarded.
The model could exploit this by learning to recall attention stored in the static bookmark (pointing at address 0) every time it processes a command marker denoting the next subsequence, which will result in overwriting the previous subsequences until the system is told to recall. This strategy is clearly more memory efficient, as the system reuses the same addresses and overwrites the memory repeatedly.

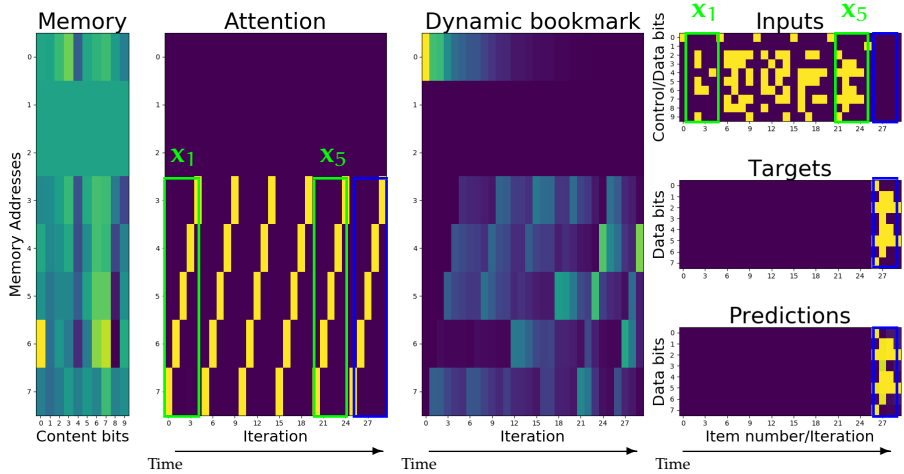
\begin{figure}[h]
 \floatbox[{\capbeside\thisfloatsetup{capbesideposition={right,center},capbesidewidth=3cm}}]{figure}[\FBwidth]
    {\begin{overpic}[width=0.73\textwidth]{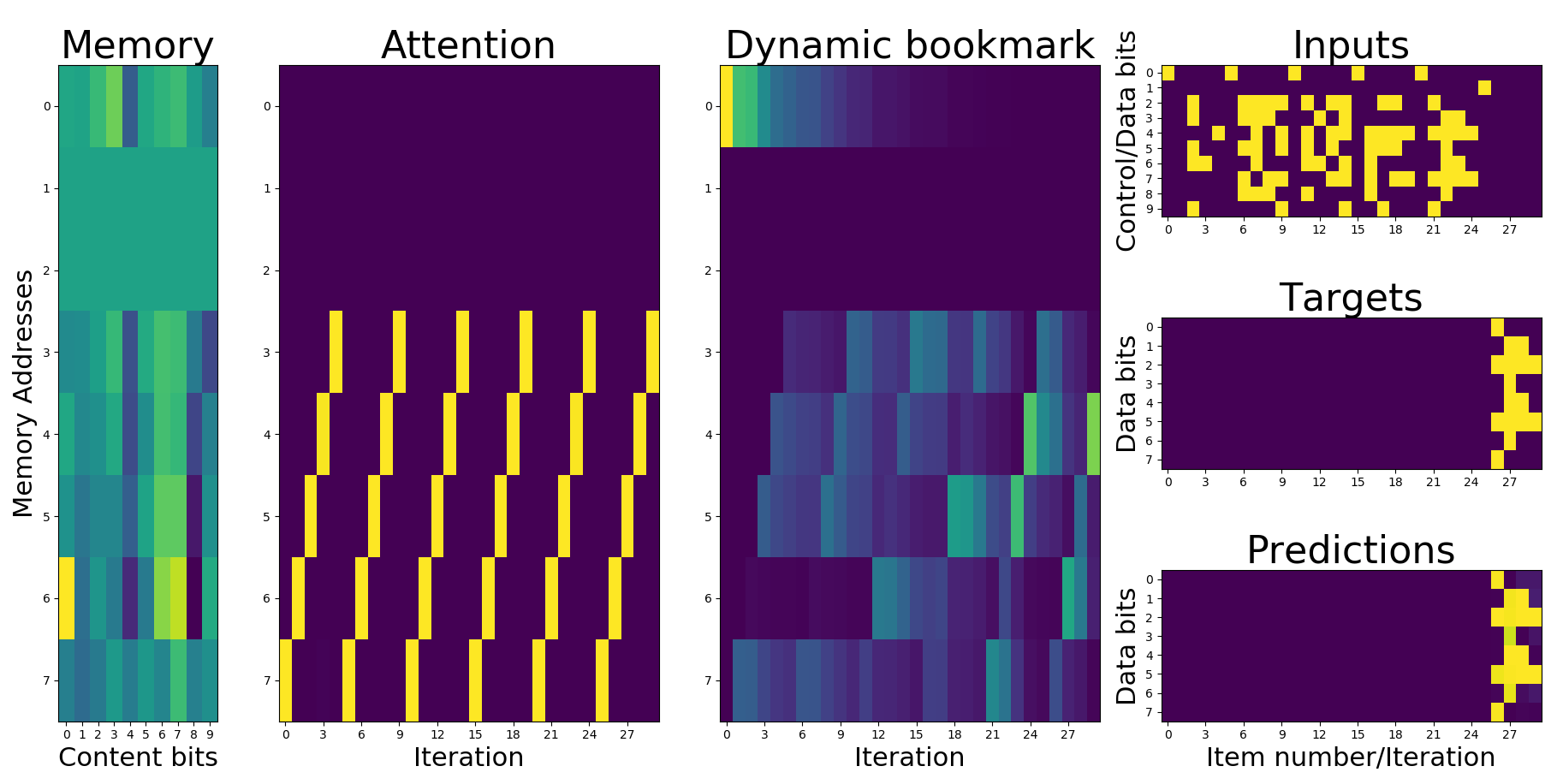}
    \linethickness{1pt}
    \put(18,4){\color{green}\framebox(3.5,26){}}
    \put(18.5,32){\color{green}{$\xx_1$}}
    \put(34.5, 4){\color{green}\framebox(3.3,26){}}
    \put(35,32){\color{green}{$\xx_5$}}
    \put(38.9, 4){\color{blue}\framebox(3.3,26){}}
    \put(75.5,36.5){\color{green}\framebox(3.3,9.5){}}
    \put(76,48){\color{green}{$\xx_1$}}
    \put(92, 36.5){\color{green}\framebox(3.3,9.5){}}
    \put(92.5,48){\color{green}{$\xx_5$}}
    \put(96.2, 36.5){\color{blue}\framebox(2.3,9.5){}}
    \put(96.2, 20.5){\color{blue}\framebox(2.3,9.5){}}
    \put(96.2, 4){\color{blue}\framebox(2.3,9.5){}}
    
    \linethickness{0.5pt}
    	\put(19,0){\color{black}\vector(1,0){23}}
	\put(16,-2){\color{black}\tiny Time}
	\put(47,0){\color{black}\vector(1,0){23}}
	\put(44,-2){\color{black}\tiny Time}
	\put(75,0){\color{black}\vector(1,0){23}}
	\put(72,-2){\color{black}\tiny Time}
	\end{overpic}}
    {\caption{"Overwrite" strategy developed by DWM for Scratch Pad (episode 969). Memory plot contains a snapshot of the memory content from the last iteration, whereas the other ones present concatenation of states from consecutive iterations (evolution in time)}
    \label{fig:scratch_pad_sufficient_memory_single_strategy_override_00969}  }

\end{figure}

To our (initial) surprise, the model \textit{always} developed the "Overwrite" strategy, irrespective of the memory size (i.e. as long as the memory size was sufficient to fit all the encoded items of a single subsequence). 
A typical example run of an early training episode is presented in \Fig{fig:scratch_pad_sufficient_memory_single_strategy_override_00969}.
Please note that memory addresses 1 and 2 remain unchanged and the model stores consecutive items of subsequences $\xx_1$ to $\xx_5$ in the same addresses 3-7.
After analyzing several runs, we hypothesize that overwriting was simpler to learn for this task because: a) both for storing and recalling the command markers, the model had to learn exactly the same behavior: recalling the attention stored in the static bookmark, b) for every other input item (data and dummy) it had to shift by one address location with the circular convolution.
As a result, it could converge rapidly by disregarding the control (update, recalling) of the dynamic bookmark (in the last training episodes the dynamic bookmark was typically "following" the current attention, despite it wasn't recalled at all).

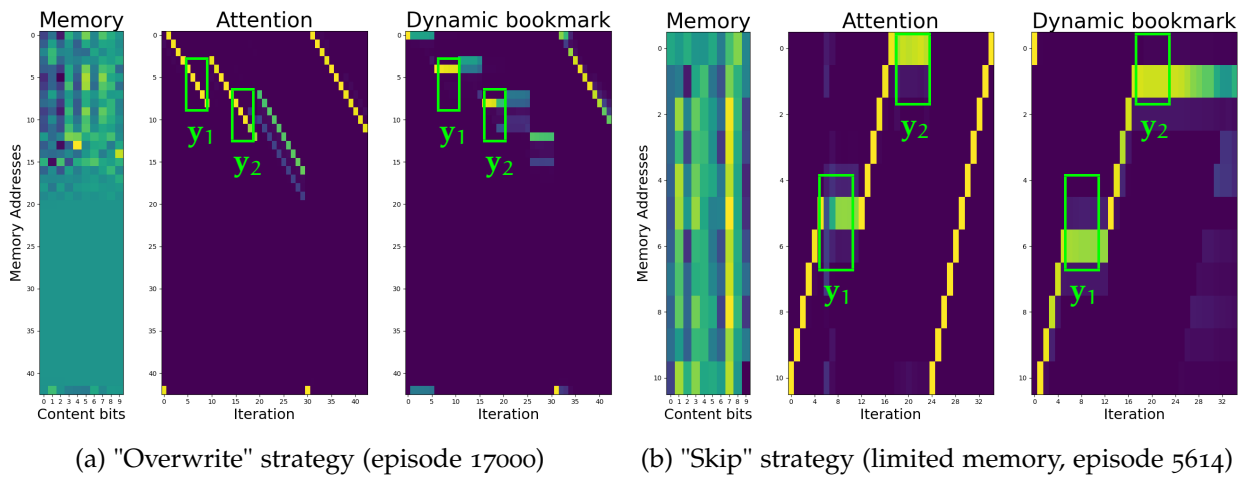
\begin{figure}[!b]
    \centering
    \null
    \begin{subfigure}[b]{0.495\textwidth}
    	\begin{overpic}[width=\textwidth]{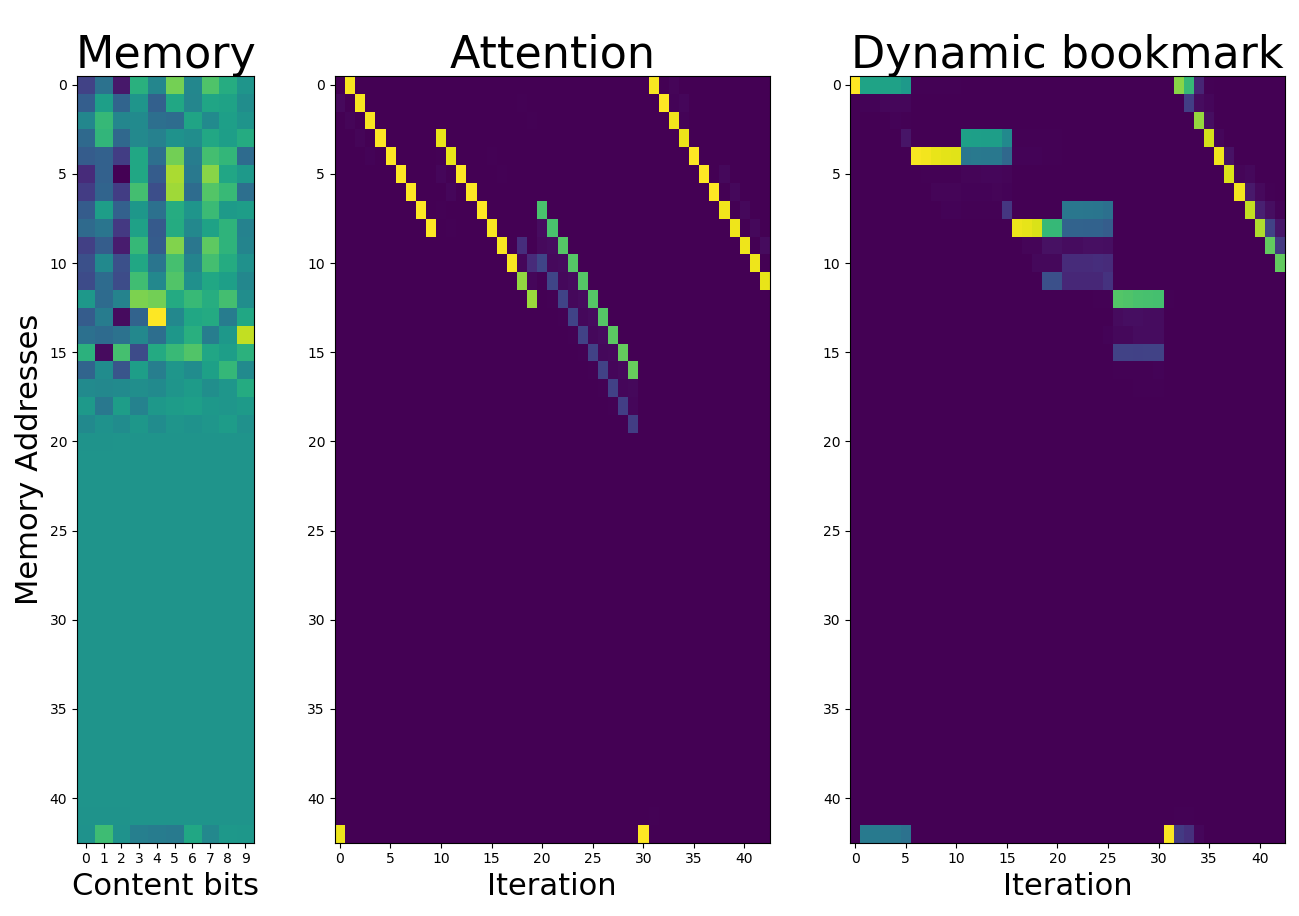}
          \linethickness{1pt}
          \put(30,52){\color{green}\framebox(3,8){}}
          \put(30,47){\color{green}{$\yy_1$}}
          \put(37.5,47){\color{green}\framebox(3,8){}}
          \put(37.5,42){\color{green}{$\yy_2$}}
          \put(71,52){\color{green}\framebox(3,8){}}
          \put(71,47){\color{green}{$\yy_1$}}
          \put(78.5,47){\color{green}\framebox(3,8){}}
          \put(78.5,42){\color{green}{$\yy_2$}}
		\end{overpic}
        \caption{"Overwrite" strategy (episode 17000)}
    		\label{fig:distraction_sufficient_memory_younes_single_strategy_override_017000}
    \end{subfigure}
    \hfill
    \begin{subfigure}[b]{0.495\textwidth}  
        \begin{overpic}[width=\textwidth]{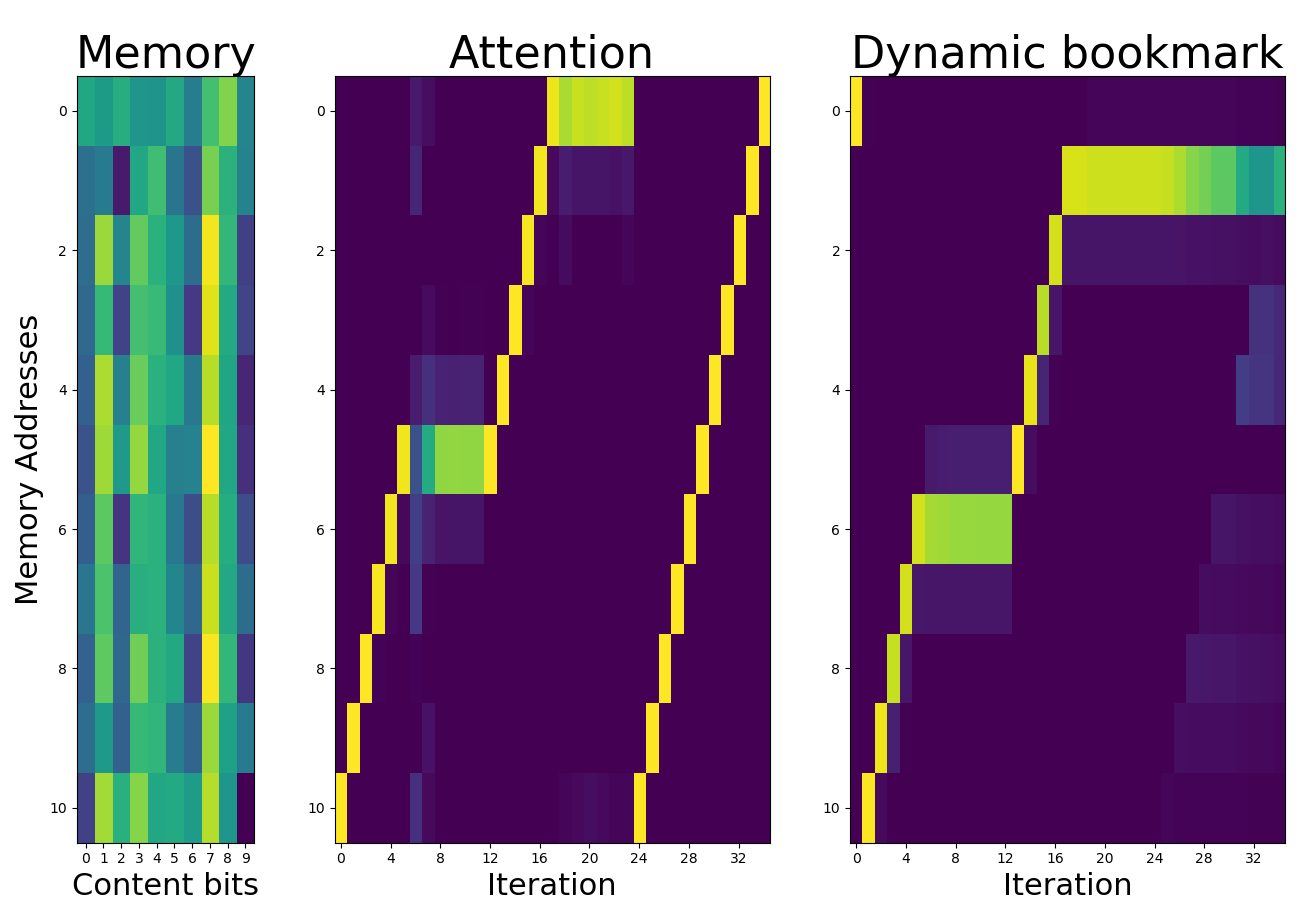}
          \linethickness{1pt}
          \put(31,26){\color{green}\framebox(5,15){}}
          \put(31.5,21){\color{green}{$\yy_1$}}
          \put(43.5,53){\color{green}\framebox(5,11){}}
          \put(44,48){\color{green}{$\yy_2$}}
          \put(71,26){\color{green}\framebox(5,15){}}
          \put(71.5,21){\color{green}{$\yy_1$}}
          \put(82.5,53){\color{green}\framebox(5,11){}}
          \put(83,48){\color{green}{$\yy_2$}}
		\end{overpic}
        \caption{"Skip" strategy (limited memory, episode 5614)}
        \label{fig:distraction_limited_memory_plus1_skip_05614}
    \end{subfigure}
    \null
    \caption{Strategies developed by DWM for solving the Ignore task (two different training runs)} 
    \label{fig:ignore_strategies_different_runs}
\end{figure}

\subsubsection{Strategies for the Ignore task}
The main goal of the Ignore task is to test the retention capabilities of the system in the presence of distractors.
For this task the input consists of two types of subsequences $\xx$ and $\yy$, where the system is supposed to ignore all $\yy_i$ and at the end recall $\xx_i$ one by one in the order of their appearance.
The task can be solved with two strategies which we call "Overwrite" and "Skip".

The "Overwrite" strategy involves overwriting of the distractors, as we have already observed during the Scratch Pad task. It assumes that system will store the consecutive items in memory and use the bookmark for moving its attention to the first address containing $\yy$ to be overwritten. The difference is, however, that in this case the model must learn to use the dynamic bookmark for that purpose.
Our experiments with sufficient memory have shown that the system can learn this strategy.  An example plot from one of the final training episodes (\Fig{fig:distraction_sufficient_memory_younes_single_strategy_override_017000})
shows that the dynamic bookmark retains its attention while processing items from $\yy_1$ and $\yy_2$. As soon as the command marker indicating $\xx$ appears, the model jumps back its attention to the dynamic bookmark and starts to overwrite.
Finally, when the recall marker appears, it recalls the attention stored in the static bookmark.

The "Skip" strategy to solve this task would be to ignore elements within the $\yy$ subsequences and \textit{skip} writing these into the memory. Our experiments with limited memory have shown that the model could also learn this strategy.
Example plots from the final episode from one of the training runs is presented in \Fig{fig:distraction_limited_memory_plus1_skip_05614}.
Please notice that in this case the model has learned to keep its attention focused on a single address for all items of $\yy_i$ and shift attention only for items belonging to $\xx_i$.

That behavior of the model that mastered the "Skip" strategy seems to be more difficult from the operational point of view.
In the "Overwrite" strategy the system develops a reactive behavior, i.e. it always performs convolutional shift except for the rare cases when it hits the command marker -- at that point it has to retrieve attention from one or the other bookmarks.
In the "Skip" strategy the command markers for $\xx$ and $\yy$ activate one of two distinct operation modes that will executed for the whole subsequence until hitting the next marker, i.e. for $\xx$ attention is supposed to move to the next address, whereas for $\yy$ it is supposed to stay at the same position.
The only way to perform this is that the controller must learn how to \textit{carry the information about the current operation mode} from one iteration to another in its hidden state, which is more difficult to learn.

\begin{figure}[!t]
    \centering
    \null
    \begin{subfigure}[b]{0.495\textwidth}
        \centering
        \includegraphics[width=\textwidth]{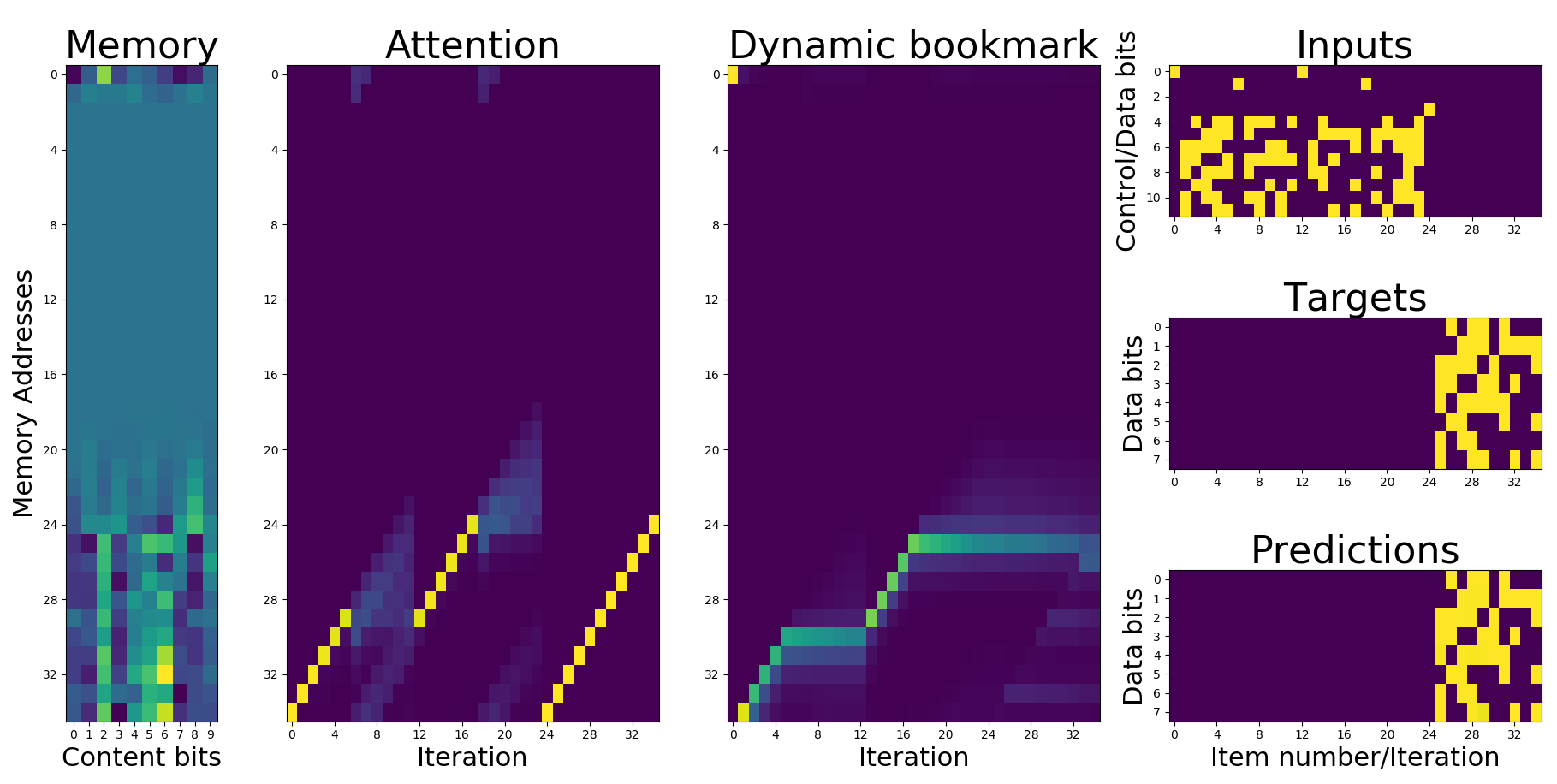}
        \caption{episode 1757: utilization of "Overwrite" strategy}
        \label{fig:distraction_sufficient_memory_change_strategy_override_01757}
    \end{subfigure}
    \hfill
    \begin{subfigure}[b]{0.495\textwidth}  
        \centering 
        \includegraphics[width=\textwidth]{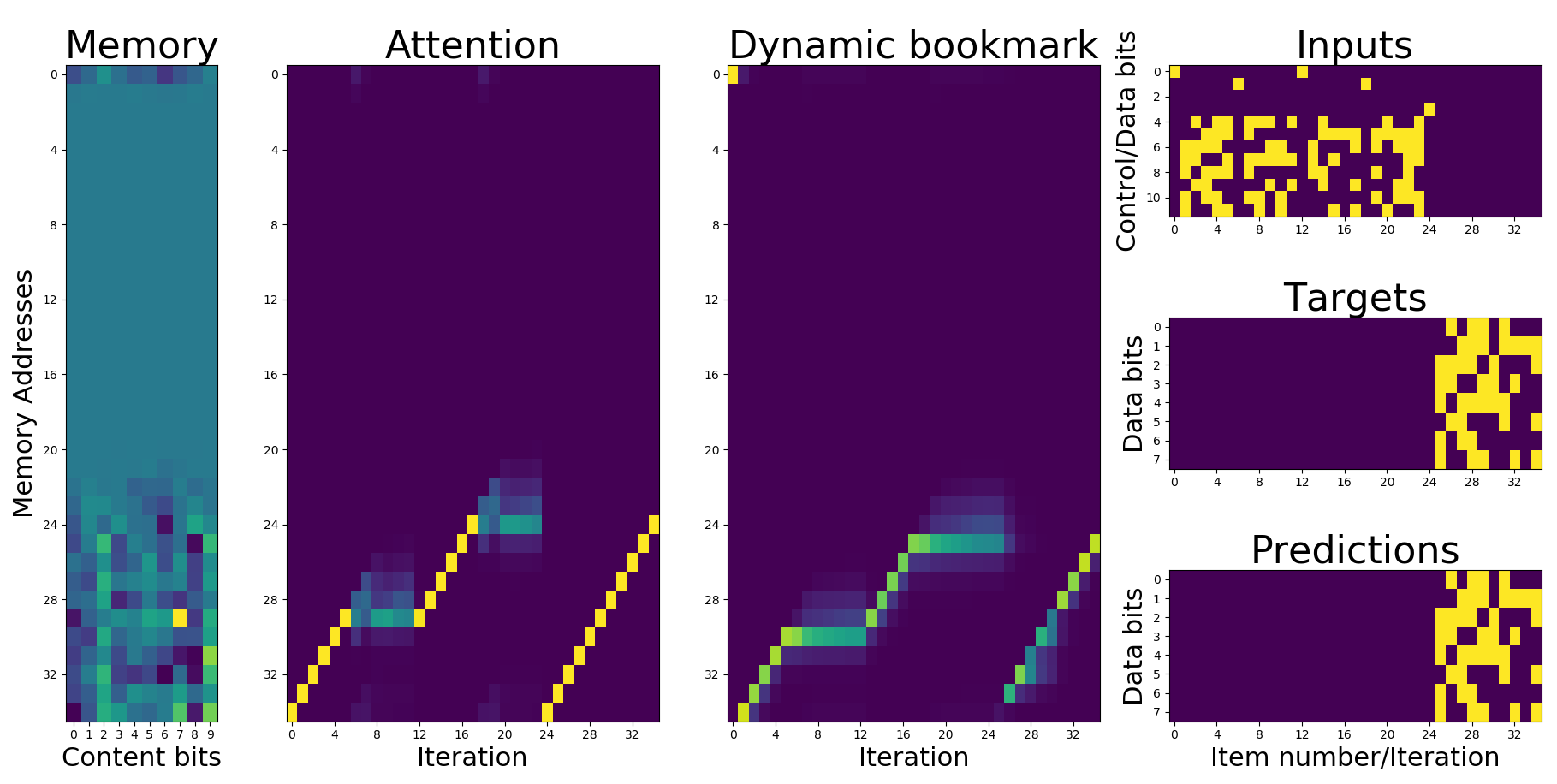}
        \caption{episode 17685: utilization of "Skip" strategy}
        \label{fig:distraction_sufficient_memory_change_strategy_skip_17685}
    \end{subfigure}
    \null
    \caption{Evolution of the strategy developed by DWM, when learning to solve the Ignore task (during a single training run, intentionally used the same sequence in both tests)} 
    \label{fig:ignore_change_strategy}
\end{figure}

We performed several experiments to support that hypothesis.
In \Fig{fig:ignore_change_strategy} we present two episodes from one of the training runs when the operation of the system seems to be evolving from one strategy to the other.
At the early stages of the training (\Fig{fig:distraction_sufficient_memory_change_strategy_override_01757}) we can observe that the shift of attention with the circular convolution is active for both types of input subsequences, whereas the dynamic bookmark already learned how to \textit{follow} attention for $\xx$ and \textit{freeze} for $\yy$.
As learning to shift attention is crucial for learning both storing and recall, model has to first master that.
However, once achieved, it seems to switch to different operation mode.
Obviously, learning two modes is simpler for dynamic bookmark, as it possesses much simpler gating mechanism and cannot shift its attention.
As the training progresses (\Fig{fig:distraction_sufficient_memory_change_strategy_skip_17685}) the model finally learns to freeze its attention when processing $\yy$ subsequences.
Developing that strategy is harder, however it more efficiently utilizes memory.

\section{Summary}

We decided to study human working memory in order to improve artificial learning models. Our analysis allowed us to distill a novel attention control mechanism (bookmarks), which is quantitatively more robust and more data-efficient than previously published work (i.e. DNC, LSTM) based on the tasks tested in this study. Moreover, our model showed generalization to sequences two orders of magnitude longer than the training regime.  We also studied the behavior of our model during learning and found out that for complex tasks, it has developed efficient strategies to control attention and use its memory resources.

In our future work we plan to use our working memory model as one component of more complex system and combine it with long-term memory and expand the capabilities to solve tasks that require both working and episodic memory.


\bibliographystyle{alpha}
\bibliography{sample}

\newcommand{\etalchar}[1]{$^{#1}$}
\begin{thebibliography}{WBCM15}

\bibitem[Bad03]{baddeley2003}
Alan Baddeley.
\newblock {Working memory: Looking back and looking forward}.
\newblock {\em Nature Reviews Neuroscience}, 4(10):829--839, 2003.

\bibitem[CE94]{conway1994working}
Andrew~RA Conway and Randall~W Engle.
\newblock Working memory and retrieval: A resource-dependent inhibition model.
\newblock {\em Journal of Experimental Psychology: General}, 123(4):354, 1994.

\bibitem[CK16]{Constantinidis2016}
Christos Constantinidis and Torkel Klingberg.
\newblock {The neuroscience of working memory capacity and training}.
\newblock {\em Nature Reviews Neuroscience}, 17:438, may 2016.

\bibitem[CKB{\etalchar{+}}05]{conway2005}
Andrew R~A Conway, Michael~J Kane, Michael~F Bunting, David~Z Hambrick, Oliver
  Wilhelm, and Randall~W Engle.
\newblock {Working memory span tasks: A methodological review and user's
  guide}.
\newblock {\em Psychonomic bulletin {\&} review}, 12(5):769--786, 2005.

\bibitem[Cow17]{cowan2017}
Nelson Cowan.
\newblock {The many faces of working memory and short-term storage}.
\newblock {\em Psychonomic Bulletin and Review}, 24(4):1158--1170, 2017.

\bibitem[CRG09]{clapp2009mechanisms}
Wesley~C Clapp, Michael~T Rubens, and Adam Gazzaley.
\newblock Mechanisms of working memory disruption by external interference.
\newblock {\em Cerebral Cortex}, 20(4):859--872, 2009.

\bibitem[DC80]{daneman1980individual}
Meredyth Daneman and Patricia~A Carpenter.
\newblock Individual differences in working memory and reading.
\newblock {\em Journal of verbal learning and verbal behavior}, 19(4):450--466,
  1980.

\bibitem[EK04]{engle2004executive}
Randall~W Engle and Michael~J Kane.
\newblock Executive attention, working memory capacity, and a two-factor theory
  of cognitive control.
\newblock {\em Psychology of learning and motivation}, 44:145--200, 2004.

\bibitem[Elm90]{elman1990finding}
Jeffrey~L Elman.
\newblock Finding structure in time.
\newblock {\em Cognitive science}, 14(2):179--211, 1990.

\bibitem[ETLC99]{engle1999}
Randall~W Engle, Stephen~W Tuholski, James~E Laughlin, and Andrew~RA Conway.
\newblock Working memory, short-term memory, and general fluid intelligence: a
  latent-variable approach.
\newblock {\em Journal of experimental psychology: General}, 128(3):309, 1999.

\bibitem[FAE02]{Fortin2002}
Norbert~J Fortin, Kara~L Agster, and Howard~B Eichenbaum.
\newblock {Critical role of the hippocampus in memory for sequences of events}.
\newblock {\em Nature Neuroscience}, 5:458, mar 2002.

\bibitem[GCB17]{gulcehre2017memory}
Caglar Gulcehre, Sarath Chandar, and Yoshua Bengio.
\newblock Memory augmented neural networks with wormhole connections.
\newblock {\em arXiv preprint arXiv:1701.08718}, 2017.

\bibitem[GCCB16]{gulcehre2016dynamic}
Caglar Gulcehre, Sarath Chandar, Kyunghyun Cho, and Yoshua Bengio.
\newblock Dynamic neural turing machine with soft and hard addressing schemes.
\newblock {\em arXiv preprint arXiv:1607.00036}, 2016.

\bibitem[GWD14]{graves14}
Alex Graves, Greg Wayne, and Ivo Danihelka.
\newblock Neural turing machines.
\newblock {\em arXiv preprint arXiv:1410.5401}, 2014.

\bibitem[GWR{\etalchar{+}}16]{graves2016hybrid}
Alex Graves, Greg Wayne, Malcolm Reynolds, Tim Harley, Ivo Danihelka, Agnieszka
  Grabska-Barwi{\'n}ska, Sergio~G{\'o}mez Colmenarejo, Edward Grefenstette,
  Tiago Ramalho, John Agapiou, et~al.
\newblock Hybrid computing using a neural network with dynamic external memory.
\newblock {\em Nature}, 538(7626):471, 2016.

\bibitem[HS97]{hochreiter1997long}
Sepp Hochreiter and J{\"u}rgen Schmidhuber.
\newblock Long short-term memory.
\newblock {\em Neural computation}, 9(8):1735--1780, 1997.

\bibitem[HT86]{hopfield1986computing}
John~J Hopfield and David~W Tank.
\newblock Computing with neural circuits: A model.
\newblock {\em Science}, 233(4764):625--633, 1986.

\bibitem[JM15]{joulin2015inferring}
Armand Joulin and Tomas Mikolov.
\newblock Inferring algorithmic patterns with stack-augmented recurrent nets.
\newblock In {\em Advances in neural information processing systems}, pages
  190--198, 2015.

\bibitem[KB14]{kingma2014adam}
Diederik~P Kingma and Jimmy Ba.
\newblock Adam: A method for stochastic optimization.
\newblock {\em arXiv preprint arXiv:1412.6980}, 2014.

\bibitem[MK08]{McNab2008}
Fiona McNab and Torkel Klingberg.
\newblock {Prefrontal cortex and basal ganglia control access to working
  memory}.
\newblock {\em Nature Neuroscience}, 11(1):103--107, 2008.

\bibitem[MRH13]{maxcey2013strategic}
Ashleigh~M Maxcey-Richard and Andrew Hollingworth.
\newblock The strategic retention of task-relevant objects in visual working
  memory.
\newblock {\em Journal of Experimental Psychology: Learning, Memory, and
  Cognition}, 39(3):760, 2013.

\bibitem[Obe09]{oberauer2009}
Klaus Oberauer.
\newblock {\em Chapter 2: Design for a Working Memory}.
\newblock Elsevier, 1st edition, 2009.

\bibitem[OL17]{oberauer2017}
Klaus Oberauer and Hsuan-yu Lin.
\newblock An interference model of visual working memory.
\newblock {\em Psychological Review}, 124(1):1--39, 2017.

\bibitem[PGC{\etalchar{+}}17]{paszke2017automatic}
Adam Paszke, Sam Gross, Soumith Chintala, Gregory Chanan, Edward Yang, Zachary
  DeVito, Zeming Lin, Alban Desmaison, Luca Antiga, and Adam Lerer.
\newblock Automatic differentiation in {PyTorch}.
\newblock 2017.

\bibitem[SBB{\etalchar{+}}16]{santoro2016meta}
Adam Santoro, Sergey Bartunov, Matthew Botvinick, Daan Wierstra, and Timothy
  Lillicrap.
\newblock Meta-learning with memory-augmented neural networks.
\newblock In {\em International conference on machine learning}, pages
  1842--1850, 2016.

\bibitem[STH18]{singh2018working}
Inder Singh, Zoran Tiganj, and Marc~W Howard.
\newblock Is working memory stored along a logarithmic timeline? converging
  evidence from neuroscience, behavior and models.
\newblock {\em Neurobiology of learning and memory}, 2018.

\bibitem[VMM05]{vogel2005neural}
Edward~K Vogel, Andrew~W McCollough, and Maro~G Machizawa.
\newblock Neural measures reveal individual differences in controlling access
  to working memory.
\newblock {\em Nature}, 438(7067):500, 2005.

\bibitem[WBCM15]{westonBCM15}
Jason Weston, Antoine Bordes, Sumit Chopra, and Tomas Mikolov.
\newblock Towards ai-complete question answering: {A} set of prerequisite toy
  tasks.
\newblock {\em CoRR}, abs/1502.05698, 2015.

\bibitem[WCB15]{weston2015memory}
Jason Weston, Sumit Chopra, and Antoine Bordes.
\newblock Memory networks.
\newblock In {\em International Conference on Learning Representations (ICLR)},
  2015.

\bibitem[Zar16]{zaremba2016learning}
Wojciech Zaremba.
\newblock {\em Learning Algorithms from Data}.
\newblock PhD thesis, New York University, 2016.

\bibitem[ZMJF16]{zaremba2016learning_icml}
Wojciech Zaremba, Tomas Mikolov, Armand Joulin, and Rob Fergus.
\newblock Learning simple algorithms from examples.
\newblock In {\em International Conference on Machine Learning}, pages
  421--429, 2016.

\end{thebibliography}

\begin{appendices}
\section{Details of operation of DWM}

\subsection{Read from the memory}
Differentiable models with addressable external memory read from the memory $M$ with $N_M$ addresses through with soft attention using the formula (e.g.~\cite{weston2015memory,gulcehre2016dynamic}):
\begin{equation}
r_t= M_t w_t,
\end{equation}
where $r_t$ is a vector read from memory.

\subsection{Memory update}
There are several schemes for memory update, e.g. using least recently used access~\cite{santoro2016meta,gulcehre2016dynamic}. In DWM we decided to use the simple erase--add scheme derived from NTM~\cite{graves14}:
\begin{equation}
M_t = M_{t-1}\circ (E-w_t \otimes e_t)+w_t\otimes a_t
\end{equation}
where E is a matrix of all ones, $e_t$ and $a_t$ are vectors of content to be erased and added to memory, respectively. $e_t$, $a_t$ and $w_t$ are emitted by interface mechanisms directed by a controller network.

\subsection{Controller}
Controller processes inputs in order to produce outputs and interface parameters. In DWM we use a single-layer recurrent neural network controller:
\begin{equation}
h_t=\sigma(W_h[x_t,h_{t-1},r_{t-1}])
\end{equation}
where $x_t$ denotes the current input and $h_{t-1}$ and $r_{t-1}$ are the hidden state and vector read from memory in the previous time step, respectively. To prevent the controller from acting as a separate working memory, the hidden state size is chosen to be smaller than that of a single input vector (in all of our experiments it was set to 5). The output logits, $y_t$ and interface vector $P_t$ are produced similarly as:
\begin{equation}
y_t=W_{y}[x_t,h_{t-1},r_{t-1}]
\end{equation}
\begin{equation}
P_t=W_{P}[x_t,h_{t-1},r_{t-1}]
\end{equation}
$W_h$, $W_y$ and $W_P$ are the only trainable parameters of our DWM model.
The interface vector $P_t$ contains all of the parameters that control reading, writing, and the attention mechanisms. Denoting the unprocessed parameters from the interface with a hat, the full list of parameters is as follows:
\begin{itemize}
\item The write vector $a_t \in \mathbb{R}^{N_M} $
\item The erase vector $e_t=\sigma(\hat{e}_t) \in [0,1]^{N_M}$ 

\item The shift vector $s_t=\softmax(\softplus(\hat{s})) \in [0,1]^3$ 
\item The bookmark update gates $g^i_t = \sigma(\hat{g}^i_t) \in [0,1]^{N_B-1}$
\item The attention update gate $\delta^i_t = \softmax(\hat{\delta}^i_t)  \in [0,1]^{N_B+1}$
\item The sharpening parameter $\gamma = 1+\softplus(\hat{\gamma}) \in [1,\infty]$
\end{itemize}

\section{Definition of input sequences}
\begin{figure*}[h!t]
    \centering
    \begin{subfigure}[b]{0.46\textwidth}
        \centering
        \includegraphics[width=\textwidth]{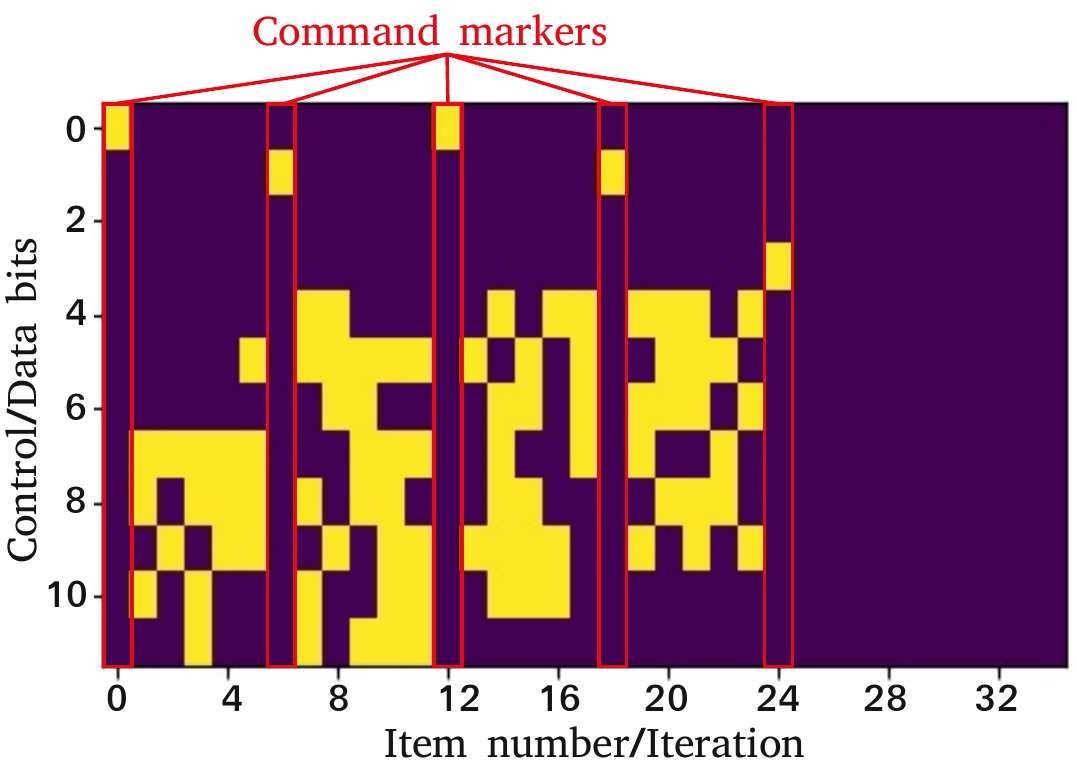}
        \caption{Command Markers} 
        \label{fig:exemplary_sequence_markers}
    \end{subfigure}
    \hspace{8mm}
    \begin{subfigure}[b]{0.46\textwidth}  
        \centering 
        \includegraphics[width=\textwidth]{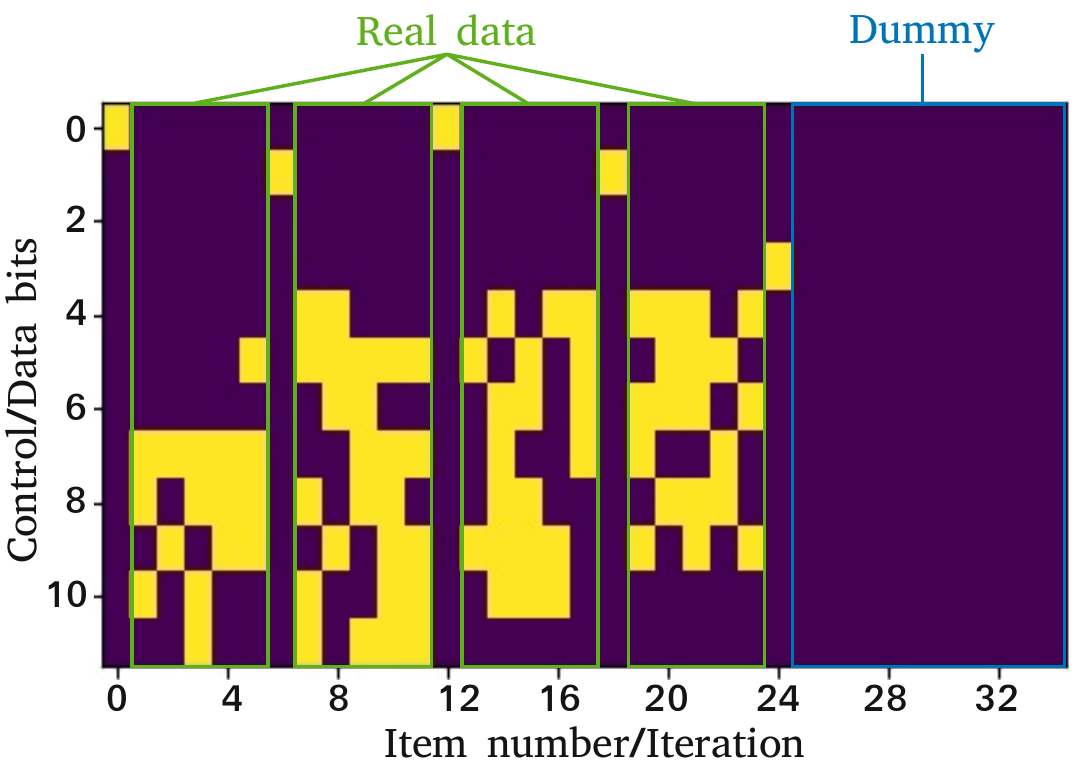}
        \caption{Real data and dummy subsequences}
        \label{fig:exemplary_sequence_data}
    \end{subfigure}
    \caption{Input sequence example} 
    \label{fig:example_sequence}
\end{figure*}

The input to every WM task is a time-indexed stream of items (\Fig{fig:example_sequence}).
At a higher level, we view the input as a  \emph{concatenation} of various subsequences that represent different functional units of processing.
For all "Simple" tasks, there is only one type of subsequence, and the  output will be reproduced from the memory with or without manipulation.
The "Complex" tasks may involve a secondary set of subsequences, which may or may not require immediate output as indicated in the "Forget and "Operation Span" tasks.

In the actual encoding of the input, we use a constant-sized set of special items (called Command Markers, please refer to \Fig{fig:exemplary_sequence_markers}) to both mark the beginning of a subsequence as well as indicate its functional type. Important note is that the system does not know a priori what kind of operation is associated with a given type of marker and must learn that from data.  We ignored markers in Table \ref{tab:tasks} to keep the description simple. 

Each subsequence is either real data or dummy (\Fig{fig:exemplary_sequence_data}).
The dummy subsequences represent elements in the input processing where a suitable target of the \emph{same} length needs to be output.
In \Cref{tab:tasks} we denote this by the $\dummy$ symbol.
Introduction of dummies enables \emph{delays} in the input processing for capturing memory retention and other aspects of WM tasks.

\begin{table}[!h]
\begin{center}
\begin{tabular}{lcccccc}
\toprule 
& \multicolumn{3}{c}{Simple tasks} & \multicolumn{3}{c}{Complex tasks} \\
\cmidrule(l){2-4} \cmidrule(l){5-7}
& Training & Validation & Testing & Training & Validation & Testing \\
\midrule
Subsequence Length 		& 1--10 & 100 & 1000 & 1--6 & 20 & 20 \\
Number of subsequences	& 1	& 1  &  1  & 1--3 & 5  & 50 \\
\bottomrule 
\end{tabular}
\end{center}
\caption{Parameters used during data generation}
\label{tab:data_generator_params_appendix}
\end{table}

Throughout the experiments, we fixed the input item size to be 8 bits supplemented by additional 2--4 control bits, depending on the task.
In order to ensure that the sequences returned by our data generators for training, validation and testing are distinctive, we followed the conditions for length and number of subsequences presented in \Cref{tab:data_generator_params_appendix}.
For the memory based models, the memory size was chosen dynamically for each episode to equal the (common) length of the input sequences within the batch.

\section{Experiments}

\subsection{Number of parameters per model}
During our experiments with LSTM model we have used stacked LSTM with 3 layers and 512 hidden units in each of them.
DNC model used an single-layer LSTM controller with 20 hidden units.
Our DWM model used RNN with sigmoid activation function and 5 hidden units.
In \Cref{tab:model_parameters} we report number of trainable parameters for each of our models.

\begin{table}[!h]
\begin{center}
\begin{tabular}{lccc}
\toprule 
& \multicolumn{3}{c}{Models} \\
\cmidrule(l){2-4}
& LSTM & DNC & DWM \\
\midrule
Number of model parameters		& 5,279,752 & 4,792 & 1,066 \\
Learning Rate 	& 5 $\cdot 10^{-3}$	& 5 $\cdot 10^{-5}$  &  1 $\cdot 10^{-2}$   \\
Optimizer 	& Adam	&  Adam &  Adam  \\
\bottomrule 
\end{tabular}
\end{center}
\caption{Number of parameters of the used models}
\label{tab:model_parameters}
\end{table}

\subsection{Number of runs and converged models}
For each model and task pair we performed 10 runs.
In \cref{tab:converged_model} we report number of models that converged, i.e. the validation loss went below the 1e-4 threshold.

\begin{table}[!h] 
\begin{center}
\begin{tabular}{lccccccccc}
\toprule
\multirow{3}{*}{Task} & \multicolumn{3}{c}{Number of Successful Runs} \\
\cmidrule(l){2-4} 
& LSTM & DNC & DWM  \\
\midrule
Serial Recall  & 0/10 & 10/10 & 10/10  \\
Reverse Recall & 0/10 & ~0/10 & ~7/10  \\
Rotate Shape   & 0/10 & ~9/10 & 10/10  \\
Reading Span   & 0/10 & ~0/10 & ~8/10  \\
Forget         & 0/10 & ~0/10 & ~5/10  \\
Operation Span & 0/10 & ~0/10 & ~3/10 \\
Scratch Pad    & 0/10 & ~1/10 & ~9/10  \\
Ignore         & 0/10 & ~0/10 & ~8/10  \\
\bottomrule
\end{tabular}
\end{center}
\caption{Success criterion: validation loss $<$ $10^{-4}$}
\label{tab:converged_model}
\end{table}

\subsection{Convergence plots}
The following plots present convergence of the best models for each of the tasks, expressed in training loss.

\begin{figure}[htbp]
    \centering
    \includegraphics[width=\textwidth]{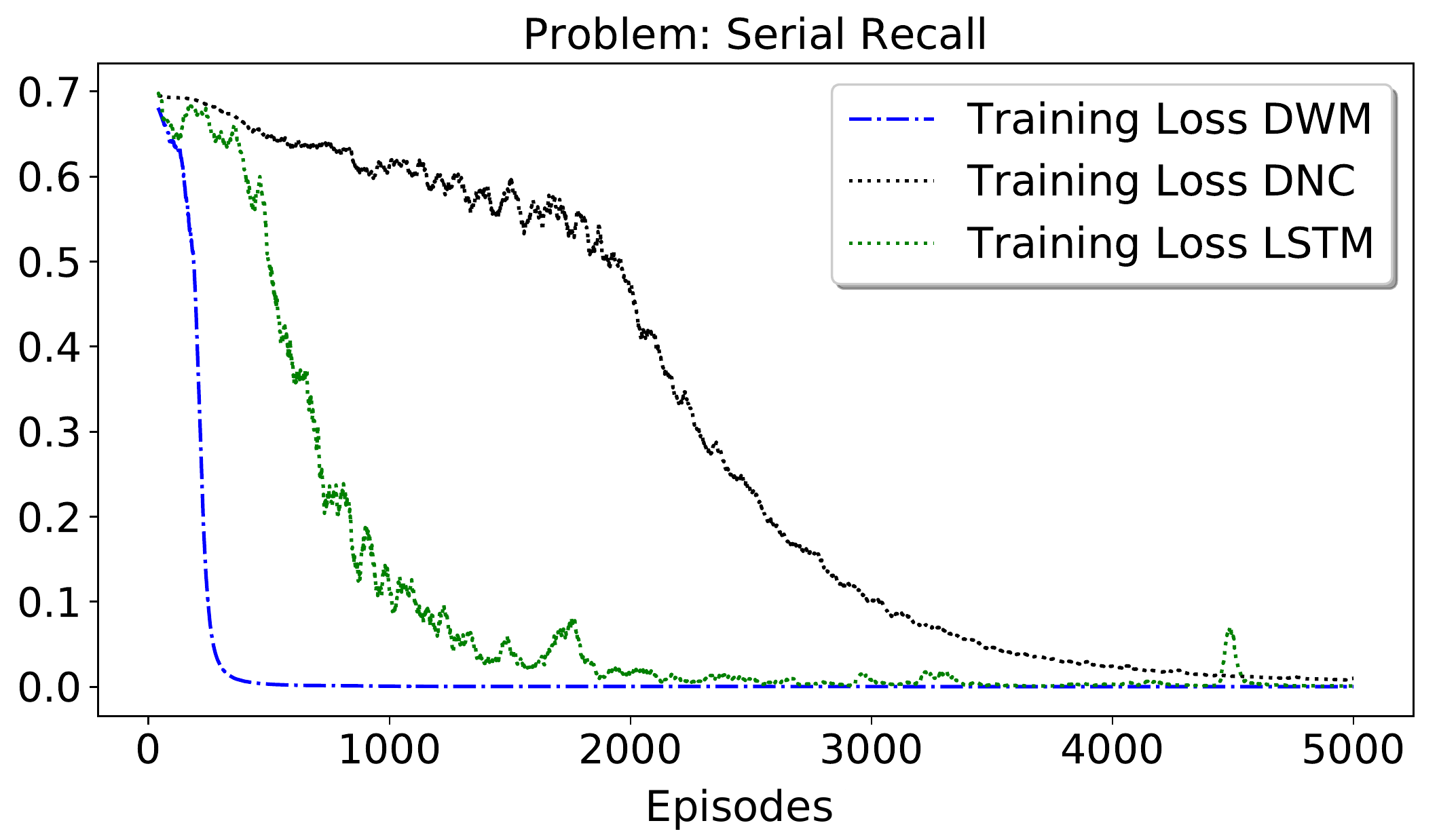}
    \caption{Training Loss of the best models on the Serial Recall task}
\end{figure}

\begin{figure}[htbp]
    \centering
    \includegraphics[width=\textwidth]{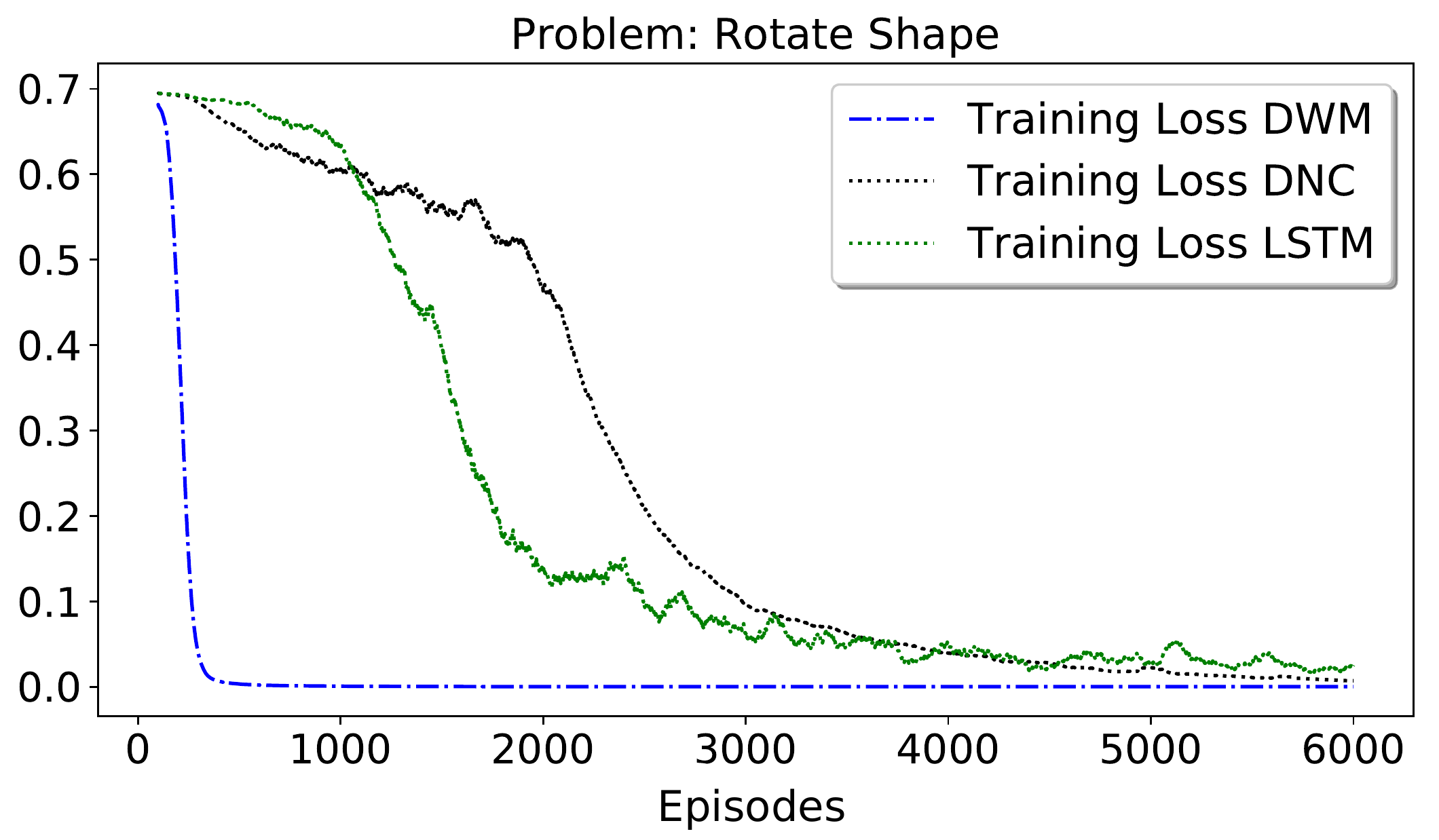}
    \caption{Training Loss of the best models on the Rotate Shape task}
\end{figure}

\begin{figure}[htbp]
    \centering
    \includegraphics[width=\textwidth]{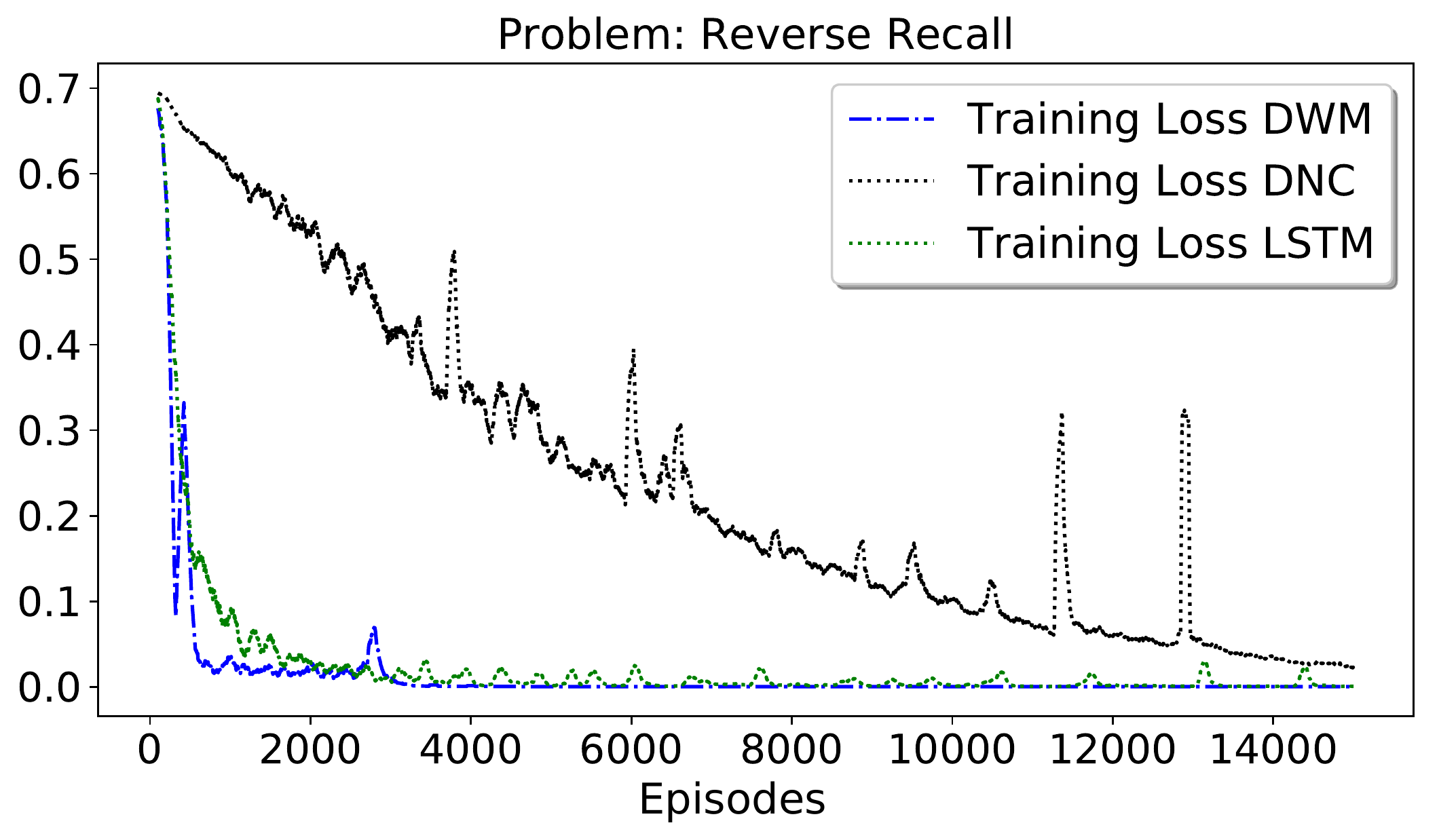}
    \caption{Training Loss of the best models on the Reverse Recall task}
\end{figure}

\begin{figure}[!b]
    \centering
    \includegraphics[width=\textwidth]{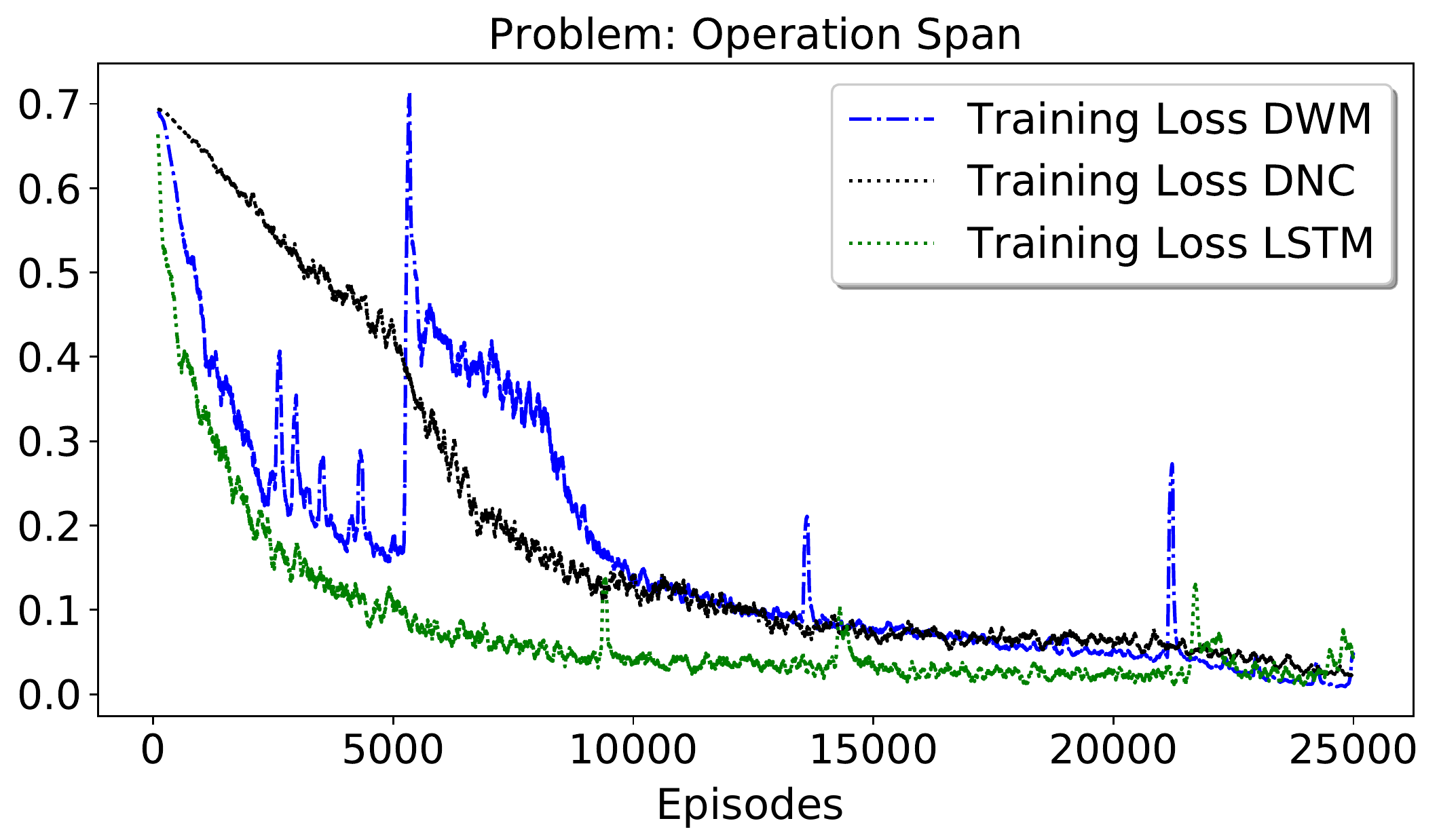}
    \caption{Training Loss of the best models on Operation Span task}
\end{figure}

\begin{figure}[!b]
    \centering
    \includegraphics[width=\textwidth]{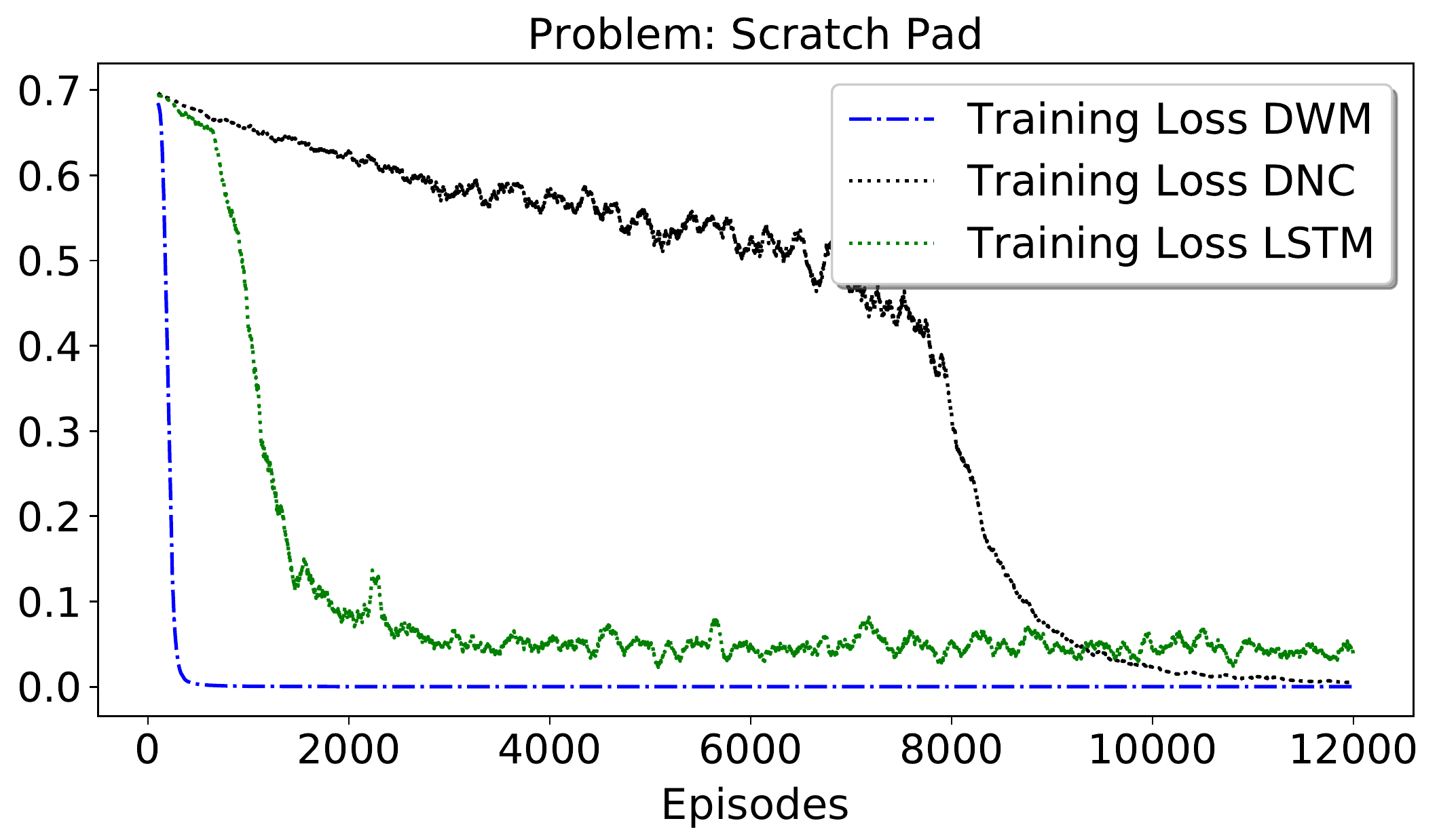}
    \caption{Training Loss of the best models on Scratch Pad task}
\end{figure}

\begin{figure}[!b]
    \centering
    \includegraphics[width=\textwidth]{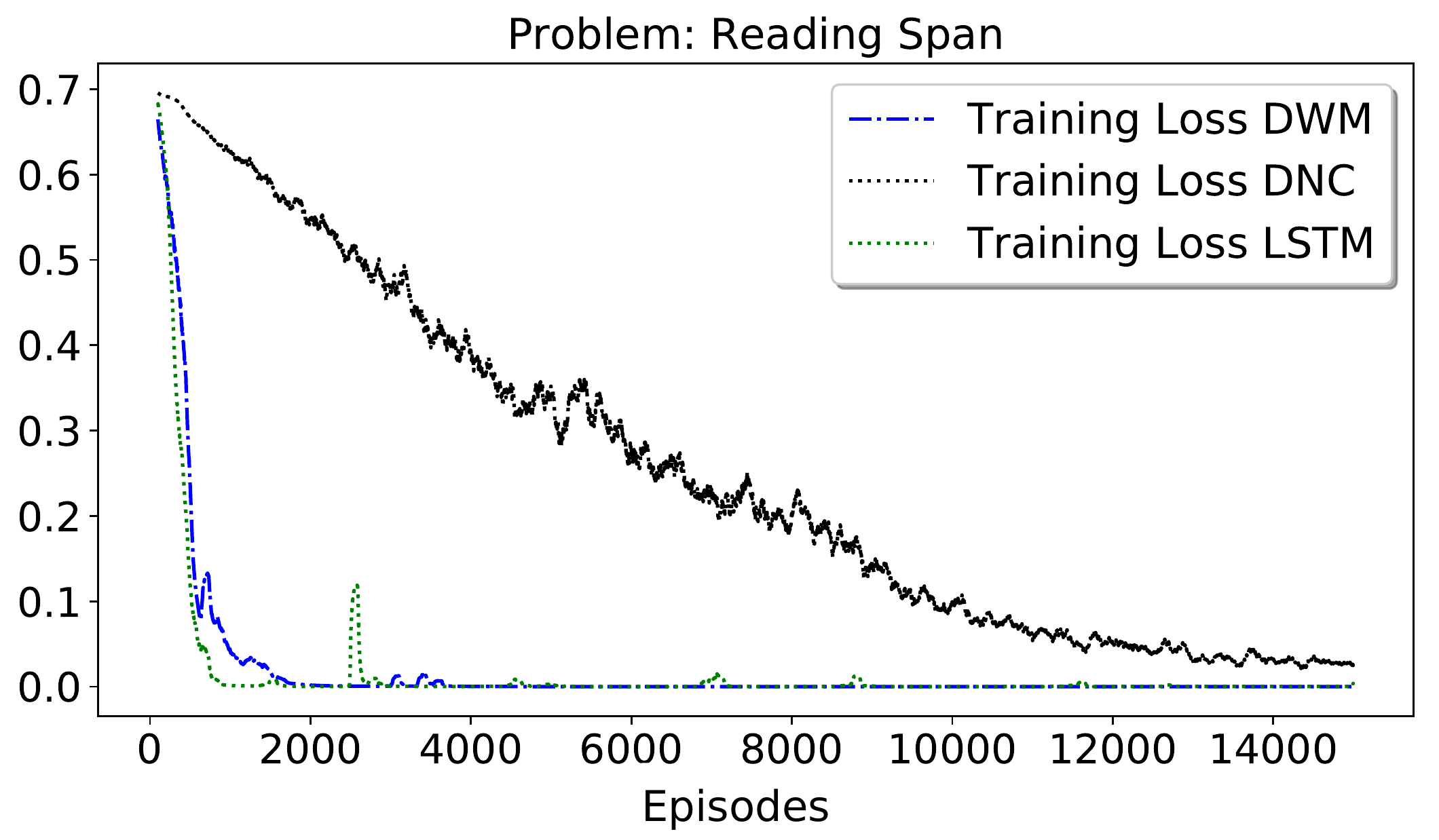}
    \caption{Training Loss of the best models on Reading Span task}
\end{figure}

\begin{figure}[!b]
    \centering
    \includegraphics[width=\textwidth]{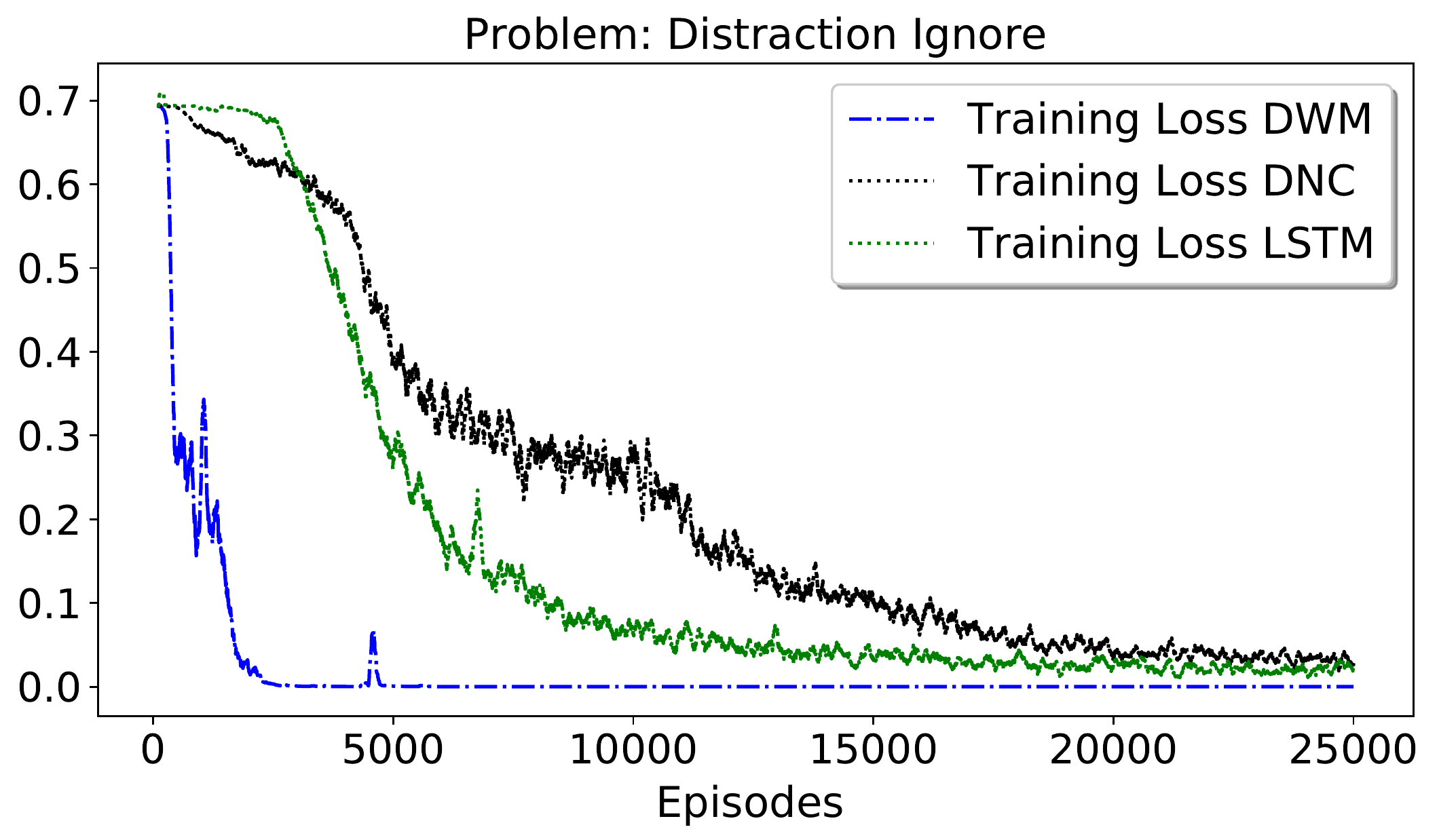}
    \caption{Training Loss of the best models on Ignore task}
\end{figure}

\begin{figure}[!b]
    \centering
    \includegraphics[width=\textwidth]{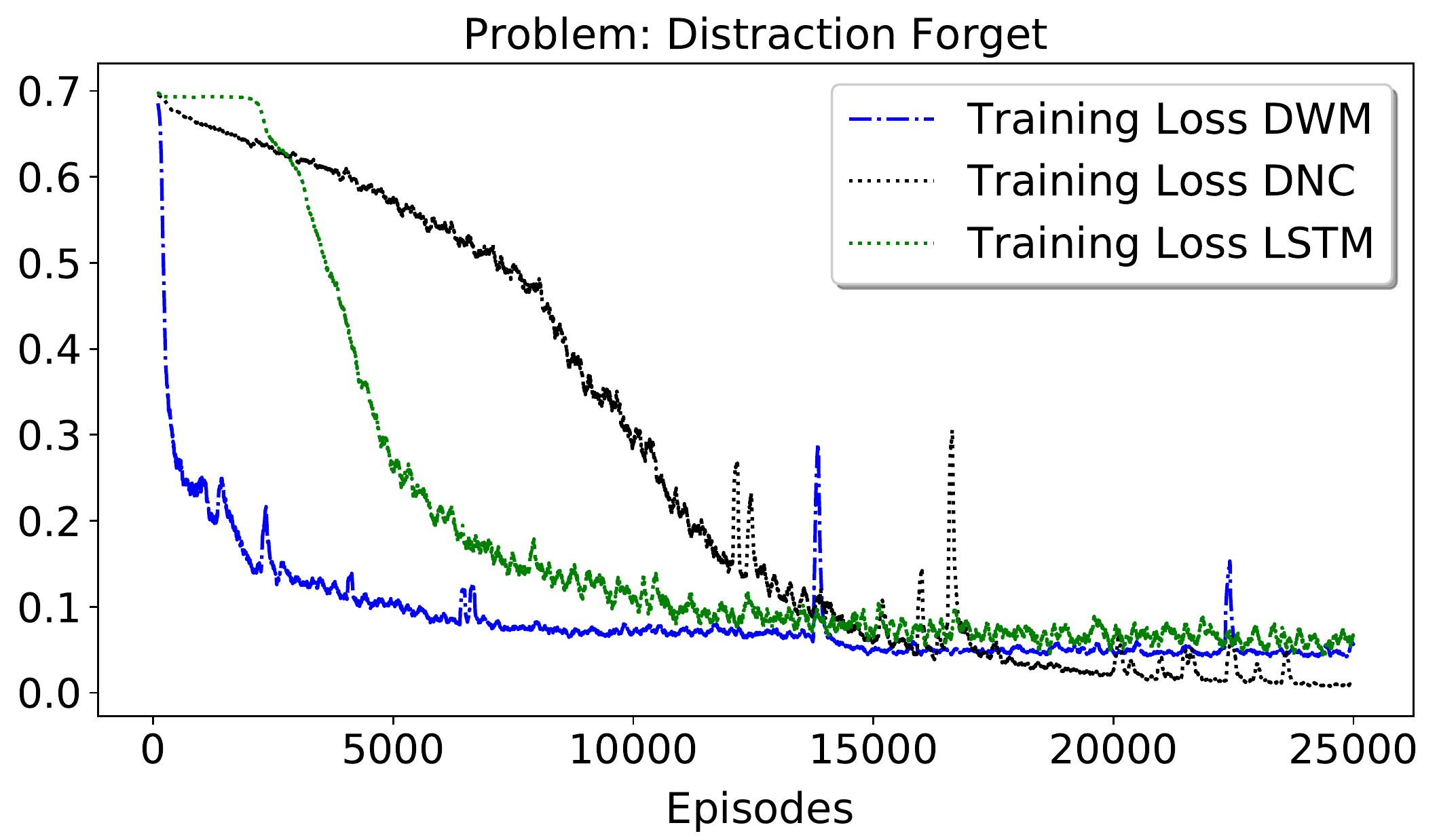}
    \caption{Training Loss of the best models on Forget task}
\end{figure}

\end{appendices}

\end{document}